\documentclass{article}

\usepackage[final, nonatbib]{neurips_ts4h_2022}

\usepackage[american]{babel}

\usepackage{mathtools} 
\usepackage{tikz} 
\usepackage{longtable}
\usepackage{hyperref}       
\hypersetup{colorlinks, linkcolor=blue}
\usepackage{multirow}
\usepackage[utf8]{inputenc} 
\usepackage[T1]{fontenc}    
\usepackage{url}            
\usepackage{booktabs}       
\usepackage{amsfonts}       
\usepackage{nicefrac}       
\usepackage{microtype}      
\usepackage{xcolor}         
 \usepackage{todonotes}
\usepackage{graphicx}
\usepackage{placeins}
\usepackage{float}
\usepackage{subfigure}

\usepackage{algorithm}
\usepackage{algorithmic}
\usepackage{amsmath,bm}
\usepackage{amssymb}

\def\Tabref#1{Table~\ref{#1}}

\def\Figref#1{Figure~\ref{#1}}

\def\eqref#1{equation~\ref{#1}}
\def\Eqref#1{Equation~\ref{#1}}

\def\1{\bm{1}}

\DeclareMathAlphabet{\mathsfit}{\encodingdefault}{\sfdefault}{m}{sl}
\SetMathAlphabet{\mathsfit}{bold}{\encodingdefault}{\sfdefault}{bx}{n}

\def\gL{{\mathcal{L}}}

\def\gU{{\mathcal{U}}}
\def\gV{{\mathcal{V}}}

\def\sR{{\mathbb{R}}}

\usepackage{multirow}
\usepackage{enumitem}
\usepackage{array}

\usepackage{amsthm}
\usepackage{wrapfig}
\usepackage{caption}

\colorlet{darkgreen}{green!40!black}
\colorlet{darkblue}{blue!75!black}
\colorlet{darkred}{red!80!black}
\definecolor{lightblue}{HTML}{0071bc}
\definecolor{lightgreen}{HTML}{39b54a}
\definecolor{pearDark}{HTML}{2980B9}

\title{Contactless Oxygen Monitoring with Gated Transformer}

\author{%
  Hao He$^{1}$\thanks{Equal contribution, order determined via random coin flip.} \quad  Yuan Yuan$^{1{\ast}}$ \quad Ying-Cong Chen$^{2{\ast}}$ \quad  Peng Cao$^1$ \quad Dina Katabi$^1$ \\  
  $^1$MIT CSAIL \quad $^2$HKUST (GZ) and HKUST\\ 
  \texttt{\{haohe, miayuan, pengcao, dk\}@mit.edu, yingcong.ian.chen@gmail.com}
}

\begin{document}

\maketitle

\begin{abstract}
With the increasing popularity of telehealth, it becomes critical to ensure that basic physiological signals can be monitored accurately at home, with minimal patient overhead. In this paper, we propose a contactless approach for monitoring patients' blood oxygen at home, simply by analyzing the radio signals in the room, without any wearable devices. We extract the patients' respiration from the radio signals that bounce off their bodies and devise a novel neural network that infers a patient's oxygen estimates from their breathing signal. Our model, called \emph{Gated BERT-UNet}, is designed to adapt to the patient's medical indices (e.g., gender, sleep stages). It has multiple predictive heads and selects the most suitable head via a gate controlled by the person's physiological indices. Extensive empirical results show that our model achieves high accuracy on both medical and radio datasets. 
\end{abstract}
\section{Introduction}
Remote health monitoring and telehealth are increasingly popular because they reduce costs and facilitate access to healthcare, particularly for people in remote locations~\cite{al2019remote}. Further, remote health monitoring can track the long-term physiological state of a patient or an older person who lives alone at home, and enable family and professional caregivers to provide timely help~\cite{celler1995remote,rf-fall,fan2020home, yang2022artificial}. Delivering such services, however, depends on the availability of solutions that continuously measure people's physiological signals at home, with minimal overhead to patients.  


Oxygen saturation is an important physiological signal  whose  at-home monitoring would benefit very old adults and individuals at high risk for low blood oxygen ~\cite{moss2005compromised}.
Oxygen saturation refers to the amount of oxygen in the blood – that is the fraction of oxygen-saturated hemoglobin relative to the total blood hemoglobin. Normal oxygen levels range from 94\% to 100\%. Lower oxygen can be dangerous, and if severe, lead to brain and lung failure~\cite{diaz2002usefulness,lapinsky1999safety}. 

Today, measuring oxygen saturation requires the person to wear a pulse oximeter on their finger, and actively measure themselves. While pulse oximeters are very helpful, they can be impractical in some at-home monitoring scenarios.  In particular, old people in their late 80’s and 90’s are at high risk for low blood oxygen~\cite{nlm}, and should regularly monitor their oxygen. Many of them however may suffer from dementia or cognitive impairment that prevent them from measuring themselves. COVID and pneumonia patients recovering at home can suffer from delirium~\cite{han2020}, which can affect their reasoning and ability to measure their oxygen levels. Additionally, blood oxygen tends to drop during sleep, making it particularly important to track oxygen overnight~\cite{palma2008oxygen,gries1996normal}. Yet people cannot actively measure themselves while asleep. 


The above use cases motivate us to try to complement pulse oximetry with a new approach that can work passively and continuously, assessing blood oxygen throughout the night without requiring the person to wear a sensor or actively measure themselves. 
%
%
Prior attempts at estimating blood oxygen passively, without a wearable sensor, rely on cameras~\cite{mathew2021remote,van2016new,van2019data,shao2015noncontact,bal2015non,guazzi2015non}. 
For example, \cite{mathew2021remote} proposes a convolution neural network that analyzes a video of the person's palm to estimate his/her blood oxygen. While not requiring wearable sensors, it cannot work continuously since the user cannot keep their hand in front of the camera for a long time. It also cannot operate in dark settings and thus cannot monitor one's oxygen during sleep. To bypass the limitations imposed by cameras, we propose a different sensing modality, \emph{radio-frequency (RF) signals}. 

We propose to monitor oxygen saturation by analyzing the radio signals that bounce off a person's body. Recent research has demonstrated the feasibility of monitoring breathing, heart rate, and even sleep stages by transmitting a very low power RF signal and analyzing its reflections off a person's body~\cite{adib2015smart,yue2018extracting}. 
The medical literature shows an inherent dependence and dynamic interaction between the breathing signal and oxygen saturation~\cite{hanson1975, na1961, Parkins1998}. Building on these advances, we use RF signals to track a person's breathing signal and train a neural network to infer oxygen from respiration.
Such a design can measure a person's oxygen without any physical contact or wearable devices. Thus, it does not burden the patient or interfere with their sleep. 
Using breathing as a mediator has an additional side benefit. It would be very hard to collect a large dataset of radio signals and the corresponding oxygen levels. Luckily however there are multiple large medical datasets that contain continuous breathing signals paired with oxygen measurements. This allows us to train a model to infer oxygen from breathing and test it directly on breathing extracted from radio signals.

While designing our model, motivated by personalized medicine, we aim for a model that can adapt to the patient's medical indices (e.g., gender, disease diagnoses). We observe that many medical indices are binary or categorical. To leverage such variables, we propose \emph{Gated BERT-UNet}, a new transformer model that has multiple predictive heads. It selects the most suitable head for each person via a gate controlled by the person's categorical indices. 
We evaluate our model on medical and RF datasets. Experiments show that our model's average absolute error in predicting oxygen saturation is 1.3\%, which is significantly lower than the state-of-the-art (SOTA) camera-based models~\cite{mathew2021remote}.

\section{Our System and Model (Gated BERT-UNet)} 


\begin{figure}[t!]
\centering
\includegraphics[width=0.75\columnwidth]{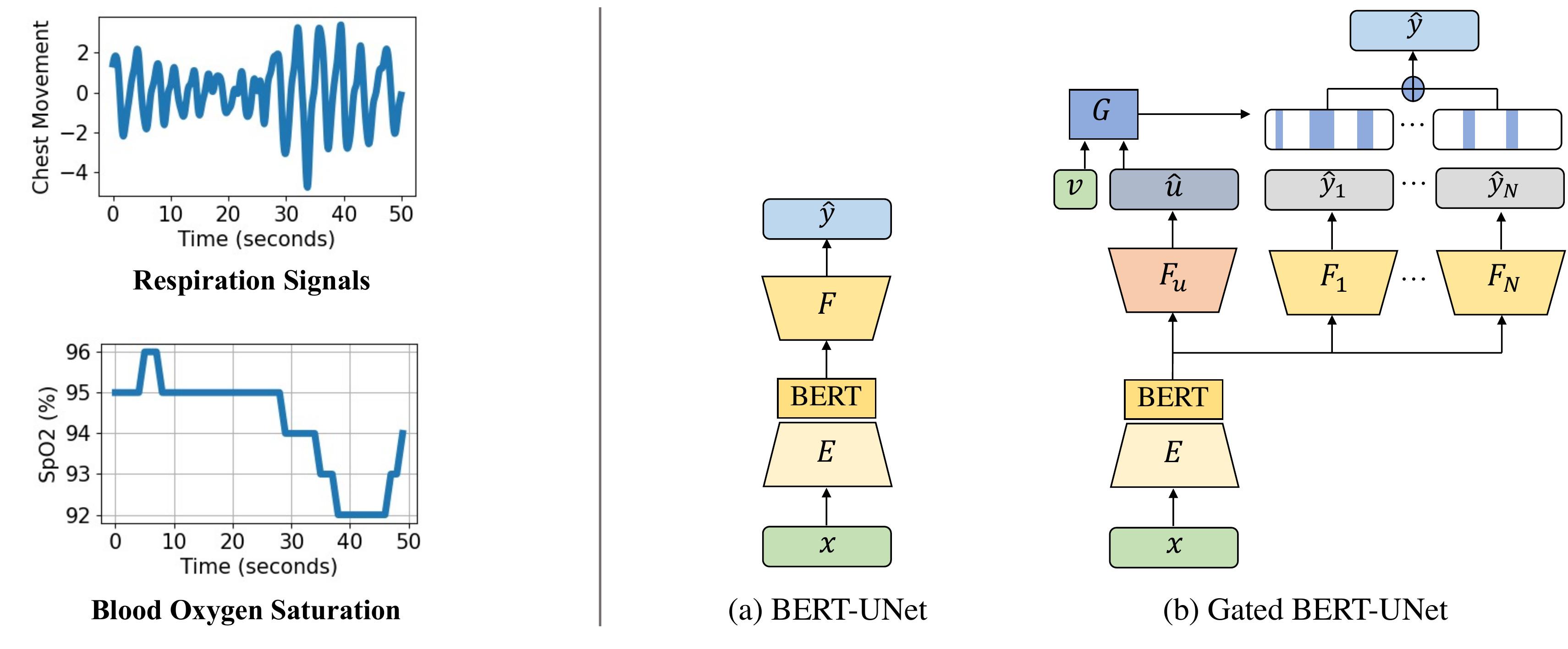}
\vspace{-1mm}
\caption{Left: Examples of respiration signals and corresponding blood oxygen saturation. Right: Illustrations of the proposed models: (a) The backbone BERT-UNet model, which contains an encoder $E$ and a predictor $F$ as the UNet structure, and a BERT module at the bottleneck of the UNet. (b) The Gated BERT-UNet, where a gate $G$, controlled by the accessible variable $v$ and the predicted inaccessible variable $\hat{u}$, is used to select among multiple predictive heads.}
\vspace{-6mm}
\label{fig:model}
\end{figure}

The proposed oxygen estimation solution operates in two steps: (1) Extracting breathing signals from RF signals and (2) Estimating SpO2 from breathing signals. Please refer to Appendix \ref{sup:br-from-rf} for the details of step one. We focus on step two in this section. 

We formulate the oxygen saturation prediction from respiration signals as a sequence-to-sequence regression task. \Figref{fig:model} left shows an example of respiration signals and corresponding blood oxygen saturation. The model takes the breathing signal $x \in \mathbb{R}^{1 \times f_{b} T}$ over a $T$-second interval and predicts the oxygen time series $y \in \mathbb{R}^{1 \times f_{o} T}$ over the same period, where $f_b$ and $f_o$ are the sampling frequencies of respiration and oxygen, respectively. In this study, the default sampling rates for respiration and oxygen are $f_b=10Hz$ and $f_o=1Hz$. 

As shown in \Figref{fig:model}(a), our backbone model is a combination of a BERT module~\cite{bert} and UNet~\cite{unet}. The \emph{BERT-UNet} backbone consists of an encoder $E$ and an oxygen predictor $F$. The encoder has convolutional layers and bidirectional attention modules while the decoder is fully convolutional and has skip-links from the encoder. The backbone is simply trained by $L_1$ loss and correlation loss. Please refer to Appendix \ref{model:basemodel} for the details of model architecture and loss functions.

Our full model, \emph{Gated BERT-UNet} augments BERT-UNet with multiple predictive heads to adapt the model's prediction to different medical indices. It is important for boosting the model's performance since the relationship between breathing and oxygen saturation has been shown related to many medical indices. For example, gender is an influential factor as men and women have differences in their oxygen transport systems~\cite{reybrouck1999gender}. Another example is that sleep stage affects a person's resting oxygen levels~\cite{choi2016severity}.

To do the adaptation, our Gated BERT-UNet selects the most suitable head for a person via a gate controlled by the subject's categorical indices. 
Specifically, as shown in \Figref{fig:model}(b), it has a gate function $G(v,u): \gV \times \gU \to \{1,2,\cdots,N\}$ where $v \in \gV$ and $u \in \gU$ are accessible/inaccessible variables, $N$ is the number of gate statuses. We use the term \emph{accessible variable} for variables easily available during inference time (e.g., gender, race) and the term \emph{inaccessible variable} for information that is not available during inference, but available during training, like a person's sleep stages.
The prediction for inaccessible variable is learned concurrently with the main task under full supervision.  

The model has $N$ heads $\{F_i\}_{i=1}^N$ that adapt the prediction $\hat{y}_i=F_i(E(x))$ to the gate status. It also has an extra predictor $F_u$ to infer inaccessible variables $u$. During testing time, based on the accessible variables $v$ and estimated inaccessible variables $\hat{u}$, we evaluate the gate status $s = G(v,\hat{u})$. Please see Appendix \ref{sec:fullmodel} for the details of (1) the loss function used to train our full model; (2) the exact construction of the gate function $G(v,u)$.


\section{Experiments and Results}
\label{sec:models}



\textbf{Dataset and Metrics.} We conduct experiments on three medical datasets: SHHS~\cite{SHHS2}, MrOS~\cite{MROS} and MESA~\cite{MESA}, and a self-collected RF dataset. We consider three evaluation metrics: correlation (Corr), mean-absolute error (MAE) and rooted mean-squared error (RMSE). Please refer to Appendix~\ref{sec:datasets} for more details on datasets and metrics. 

\textbf{Baselines.} We compare our model with the following baselines: (a)~\textit{CNN}, a fully convolutional model; (b)~\textit{CNN-RNN}, an augmentation of CNN with RNN units in the bottleneck; (c) \textit{BERT-UNet + VarAug}, which takes BERT-UNet as the backbone and uses medical variables as extra inputs or outputs. Appendix \ref{sec:baseline} explains more details of the baselines.

\textbf{Training and Evaluation Protocols.}  Since the RF dataset is too small for training, we train models on breathing signals from the medical datasets, and test them directly on the respiration signals extracted from the RF dataset as well as medical datasets' test sets. More details are in Appendix \ref{sec:protocol}.

\subsection{Results on Medical Datasets}

\begin{table*}[]
\small
\centering
\caption{Performances on medical datasets. (* indicates using extra physiological variables.)}
\label{tab:res-medical}
\resizebox{0.9\textwidth}{!}{

\begin{tabular}{c||p{5mm}p{5mm}p{8mm}|p{5mm}p{5mm}p{8mm}|p{5mm}p{5mm}p{8mm}||p{5mm}p{5mm}p{8mm}}
\toprule
\multirow{2}{*}{Model}   & \multicolumn{3}{c|}{SHHS}                     & \multicolumn{3}{c|}{MESA}                     & \multicolumn{3}{c||}{MrOS}                     & \multicolumn{3}{c}{Overall}                  \\ 
& Corr$^\uparrow$ & MAE$^\downarrow$ & RMSE$^\downarrow$ & Corr$^\uparrow$ & MAE$^\downarrow$ & RMSE$^\downarrow$ & Corr$^\uparrow$ & MAE$^\downarrow$ & RMSE$^\downarrow$ & Corr$^\uparrow$ & MAE$^\downarrow$ & RMSE$^\downarrow$ \\ \hline 
CNN   &  0.47 & 1.72 &  1.81 & 0.46  & 1.62  & 1.72 & 0.49 & 1.77 &1.82 & 0.47  & 1.73 & 1.78  \\
CNN-RNN                  & 0.48          & 1.70          & 1.80          & 0.50          & 1.56          & 1.65          & 0.52          & 1.76          & 1.84          & 0.49          & 1.67          &   1.77           \\
BERT-UNet                & 0.51          & 1.67          & 1.77          & 0.52          & 1.55          & 1.66          & 0.54          & 1.75          & 1.84          & 0.51          & 1.64          & 1.76          \\ \hline 
BERT-UNet + VarAug*       & \textbf{0.52}          & 1.65          & 1.76          & 0.52          & 1.51          & 1.62          & \textbf{0.55}          & 1.68          & 1.78          & 0.52          & 1.61          & 1.72          \\
Gated BERT-UNet* & \textbf{0.52} & \textbf{1.61} & \textbf{1.72} & \textbf{0.53} & \textbf{1.50} & \textbf{1.61} & \textbf{0.55} & \textbf{1.65} & \textbf{1.75} & \textbf{0.53} & \textbf{1.58} & \textbf{1.70} \\ 
\bottomrule
\end{tabular}
}
\vspace{-5mm}

\end{table*}

\begin{figure*}[t!]
\centering
\includegraphics[width=0.7\textwidth]{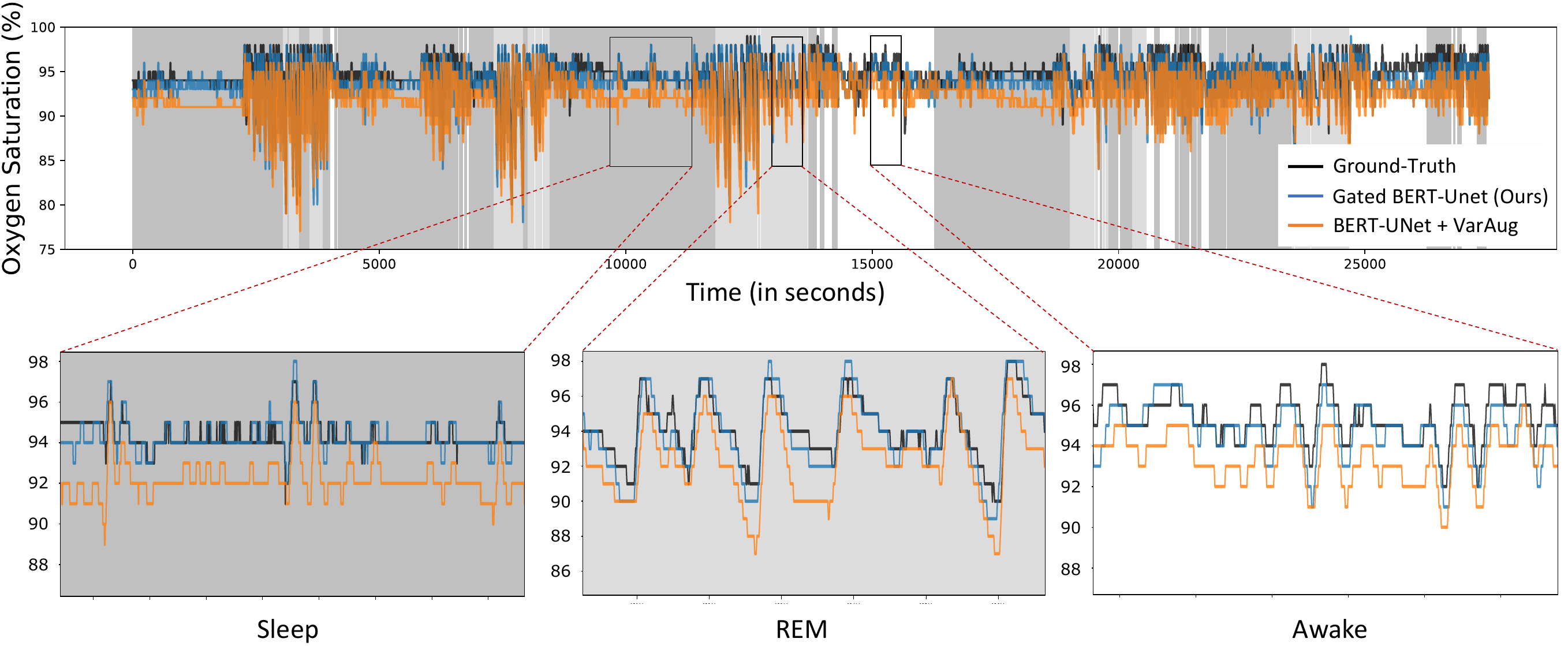}
\vspace{-2mm}
\caption{An illustrative example of the oxygen level predicted by the BERT-UNet + VarAug (Orange Curve), and our Gated BERT-UNet model (Blue Curve). The background color indicates sleep stages. The `dark grey', `light grey' and `white' corresponds to non-REM `Sleep', `REM' and `Awake'.}
\vspace{-6mm}
\label{fig:visualize_shhs2}
\end{figure*}

\noindent \textbf{Quantitative.}
The results on medical datasets are shown in \Tabref{tab:res-medical}. 
Since there is no past work that predicts oxygen from breathing or radio signals, all models in the table refer to variants of our neural network. 
The table shows that all variants achieve relatively low prediction errors with an average MAE that ranges from 1.58 to 1.73 percent, and an average RMSE that ranges from 1.70 to 1.78 percent. Such a relatively low RMSE shows our model rarely has predictions that largely deviate from the ground truth. All variants also achieve reasonable high correlations ranging from 0.47 to 0.53. Such a correlation level indicates our model's prediction capture the dynamics of the ground truth SpO2 which is also visually shown in \Figref{fig:visualize_shhs2}.
These quantitative results highlight that our system can be useful for continuous monitoring of patients' oxygen at home. 

The upper rows in the table show the variants that do not leverage side variables. The table shows that the BERT-UNet model consistently outperforms the CNN model and the CNN-RNN model on all datasets, in all metrics. This indicates that BERT-UNet is a preferable architecture for this task.

The bottom rows in the table show the results of models leveraging accessible and inaccessible medical variables.  
The table shows that {BERT-UNet + VarAug} and {Gated BERT-UNet}  outperform models that do not leverage physiological indices, demonstrating the benefit of leveraging such variables.
In addition, {Gated BERT-UNet} outperforms {VarAug} on all three datasets, demonstrating that a gated multi-head approach works best for such categorical side variables.

\noindent \textbf{Qualitative.}
Figure~\ref{fig:visualize_shhs2} visualizes the predicted oxygen saturation of the Gated BERT-UNet model and the VarAug model on a male subject in the SHHS dataset. As the ground-truth oxygen saturation are integers, we round the predicted oxygen values. The background color indicates different sleep stages. The `dark grey', `light grey' and `white' correspond to non-REM `Sleep', `REM' and `Awake' stage, respectively. We observe that Gated BERT-UNet consistently outperforms VarAug over the whole night. The small panel focuses on different sleep stages. In general, different sleep stages tend to show different behavior and hence the importance of using a gated model. Specifically, oxygen is typically more stable during non-REM `Sleep' than during REM and Awake stages.  The figures show that Gated BERT-UNet can track the ups and downs in oxygen and is significantly more accurate than VarAug. This experiment demonstrates that the way we incorporate the sleep stages into the model improves performance across different sleep stages.





\subsection{Results on RF Dataset}

%
The results on the RF dataset are shown in \Tabref{tab:umass}. All model variants have low prediction error and high correlation. Among the models that do not leverage physiological variables, {CNN-RNN} and {Bert-UNet} perform better than the vanilla {CNN}, which shows the importance of modeling temporal information. When using physiological variables, the performance of {BERT-UNet+VarAug} is similar to that of {BERT-UNet}, while {Gated BERT-UNet} is better than all other variants. This demonstrates that the gating design leverages auxiliary variables better. Overall, the MAEs, RMSEs and correlations on the RF dataset are comparable to those on the medical datasets. This indicates that our model is directly applicable to respiration signals from RF. 
We have also visualized the prediction results in \Figref{fig:visualize_rf}. As shown, our model can accurately track the fluctuation of ground truth SpO2. 

\begin{table}
	\begin{minipage}{0.42\linewidth}
		\caption{Results on the RF dataset.}
		\label{tab:umass}
		\centering
		\resizebox{0.95\textwidth}{!}{
        \begin{tabular}{c||p{5mm}p{5mm}m{9mm}}
        \toprule
        Model & Corr$^\uparrow$  & MAE$^\downarrow$ & RMSE$^\downarrow$\\
        \hline
        CNN    & 0.45 & 1.65 & 1.73    \\ 
        CNN-RNN  & 0.49 & 1.85 & 1.93  \\
        BERT-UNet  & 0.48 &  1.49 & 1.58 \\ 
        \hline
        BERT-UNet + VarAug & 0.49 & 1.49 & 1.59 \\
        Gated BERT-UNet & \textbf{0.52} & \textbf{1.32} & \textbf{1.54}\\ 
        \bottomrule
        \end{tabular}
        }
	\end{minipage}\hfill
	\begin{minipage}{0.6\linewidth}
		\centering
		\includegraphics[width=0.9\textwidth]{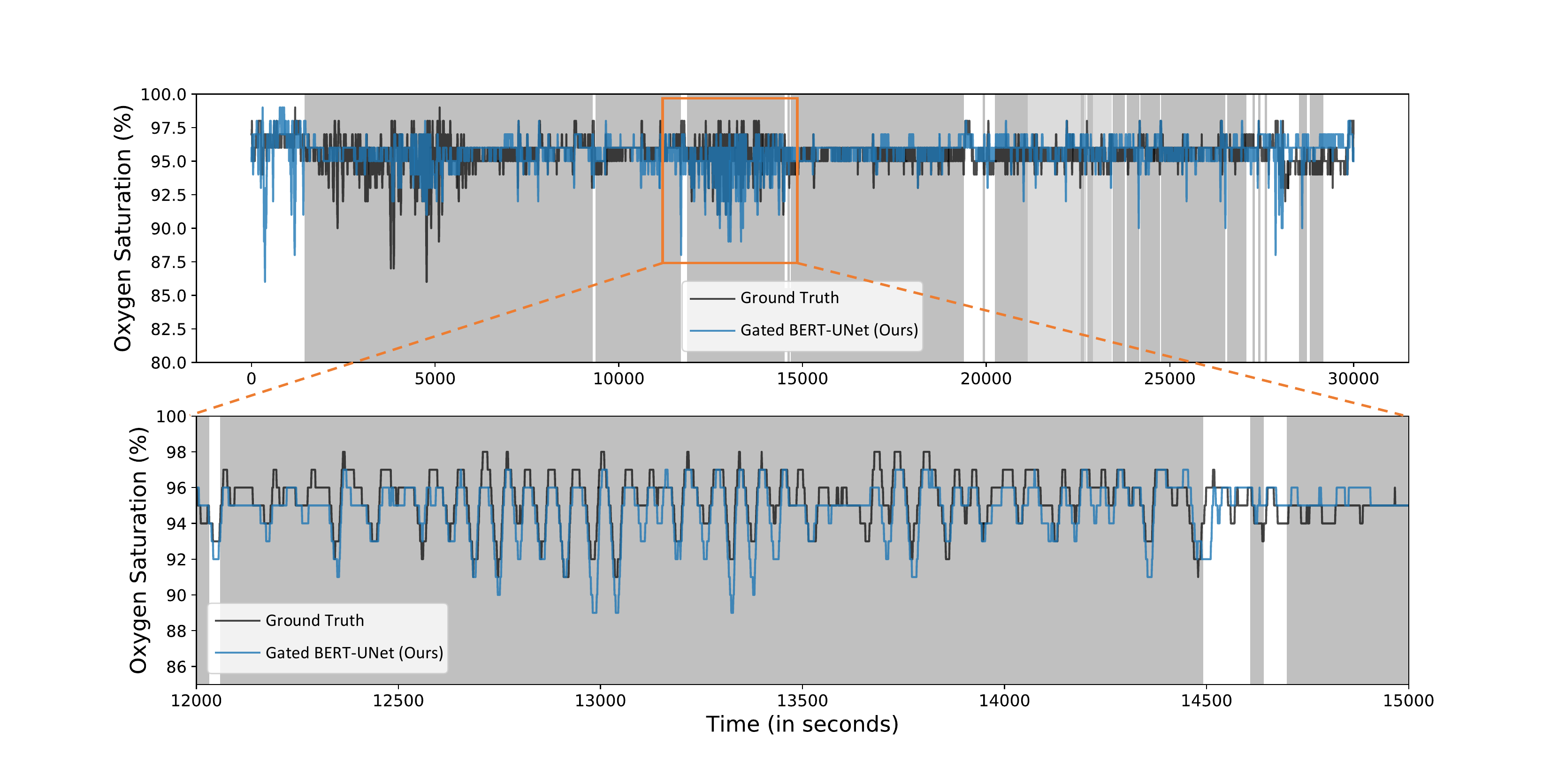}
		\captionof{figure}{Example of SpO2 prediction results on RF dataset.}
		\label{fig:visualize_rf}
	\end{minipage}
	\vspace{-10mm}
\end{table}

\subsection{Visualization of Breathing-Oxygen Patterns}\label{sec:diffpattern}
\begin{figure*}[t!]
\centering
\begin{tabular}{@{\hspace{0mm}}c@{\hspace{0mm}}c@{\hspace{0mm}}c@{\hspace{0mm}}c}
\includegraphics[width=0.33\columnwidth]{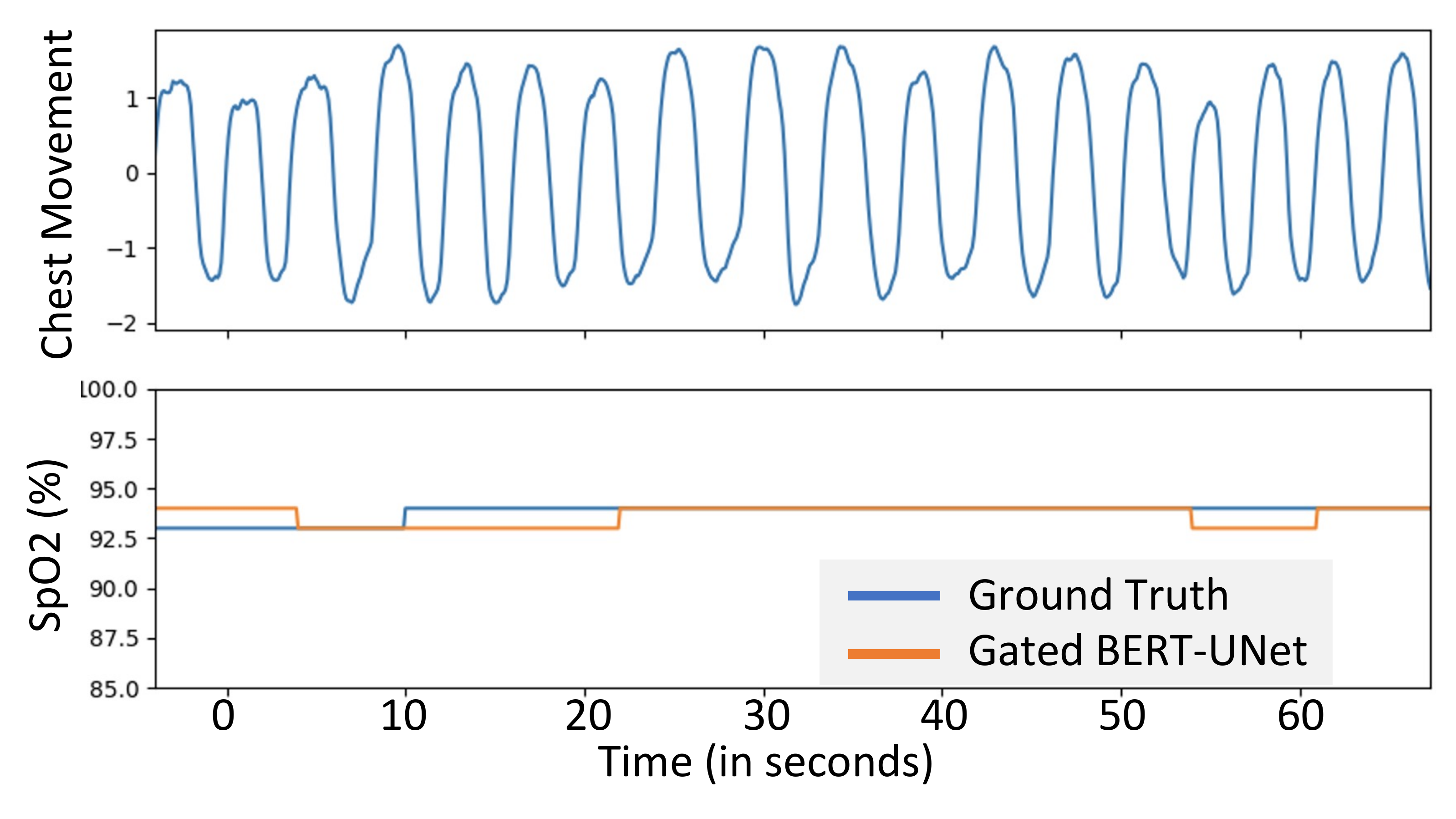}\vspace{-2mm}
 &
\includegraphics[width=0.315\columnwidth]{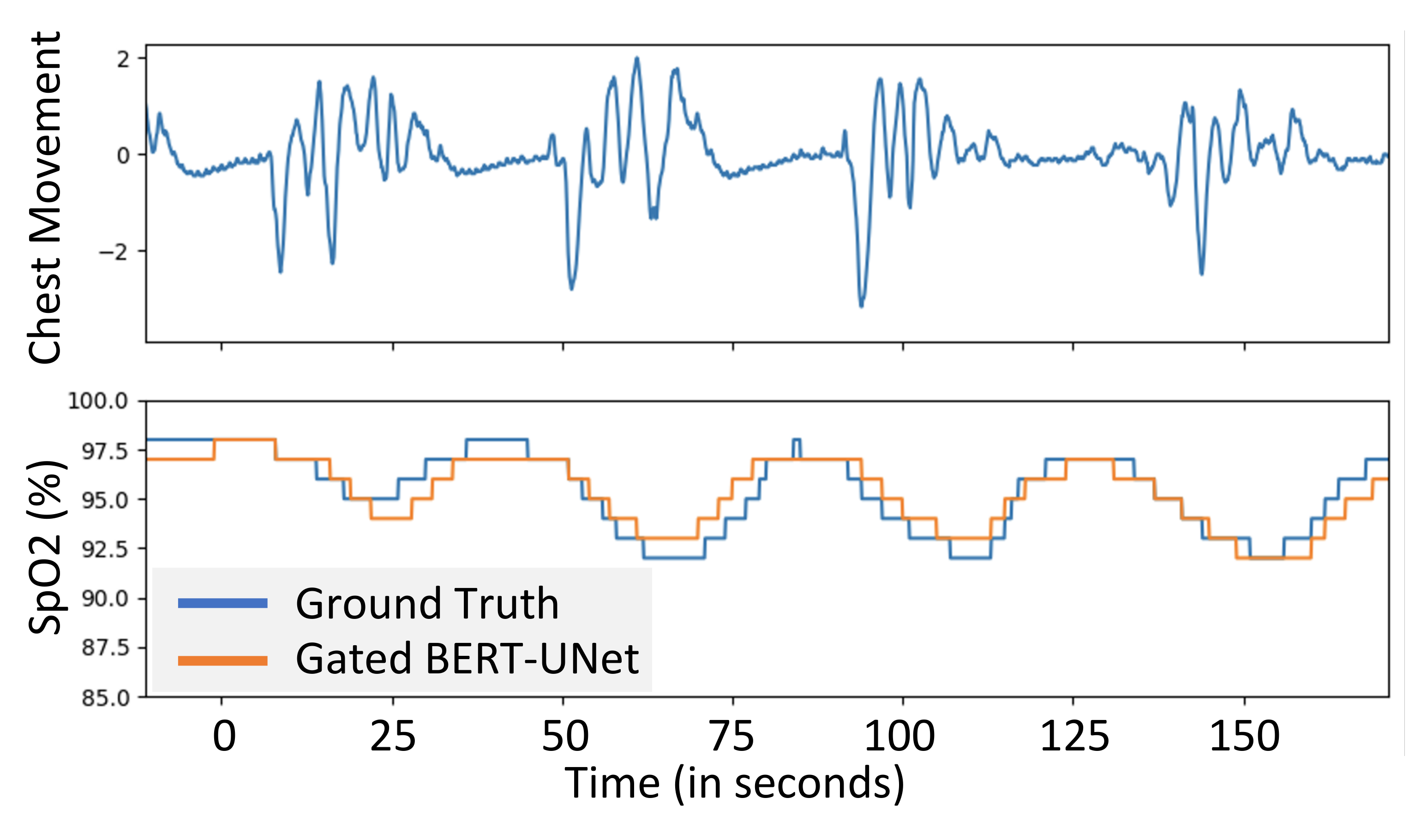} &
\includegraphics[width=0.33\columnwidth]{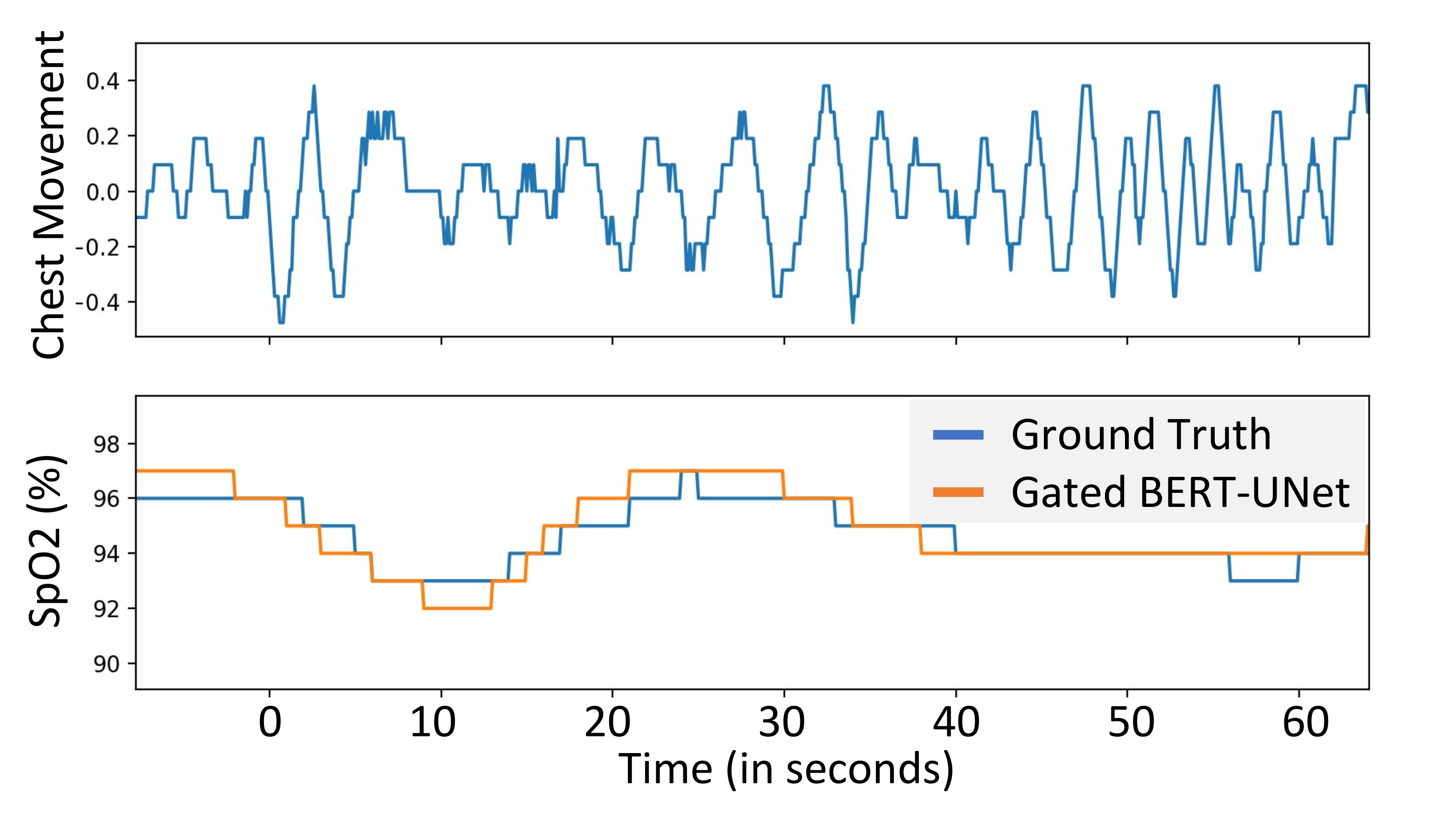} \\
(a) & (b) & (c)
\end{tabular}
\caption{
Examples of breathing signals and corresponding predicted oxygen saturation. (a) is an example of normal breathing signals. (b)(c) are examples of different abnormal breathing signals. In each figure, the first row shows the breathing signals while the second row shows the ground truth oxygen saturation (in blue) and the oxygen saturation predicted by our model (in orange).}\label{fig:breathing}
\end{figure*}

We demonstrate several visual results on different patterns of breathing signals and the corresponding ground truth/predicted oxygen saturation, as shown in Figure~\ref{fig:breathing}. \Figref{fig:breathing}(a) shows a normal breathing pattern, which leads to constant oxygen saturation. In contrast, \Figref{fig:breathing}(b,c) present two different abnormal breathing signals and the resulting fluctuated oxygen predictions.  These figures show the diversity of the oxygen and breathing patterns as well as the complexity of their relationship. The model however is able to capture this relationship for highly diverse patterns.

\subsection{Comparison with Past Works}\label{sec:cmppast}

We compare our approach with two recent deep-learning camera-based SpO2 monitoring methods.  The first method~\cite{ding2018measuring} asks the user to press his/her finger against a smartphone camera, and uses a CNN to estimate SpO2 from the video. The second method~\cite{mathew2021remote} uses a CNN to estimate oxygen from a video of the person's hand. The input of these systems is different from ours (camera vs. RF); so to compare them we follow the setup in~\cite{mathew2021remote}. 
In their setting, test samples are 180-240 seconds and vary between normal breathing to no or minimal breathing. Similarly, we divide the RF test dataset into non-overlapping 240-second segments containing both regular breathing and shallow breathing (i.e., apnea or hypopnea) and compute metrics on them. 
Table~\ref{tab:compare_w_previous_work} reports the output of the three models. The results of the baselines are taken from~\cite{mathew2021remote}, and the results of our RF-based model are computed as described. 
The results show that our model improves the correlation and reduces the MAE and RMSE in comparison to past work.



\begin{table}[t!]
\footnotesize
\centering
\caption{Prior SOTA oxygen saturation prediction methods~\cite{ding2018measuring,mathew2021remote} v.s. our \textit{Gated BERT-UNet}.}
\label{tab:compare_w_previous_work}
\begin{tabular}{c||p{5mm}p{5mm}p{8mm}|c}
\toprule
Method & Corr$^\uparrow$   &  MAE$^\downarrow$ & RMSE$^\downarrow$ & Contactless    \\ \hline
Ding et al. 2018  & 0.26 &  2.43 & 2.85 & No  \\ \hline
Mathew et al. 2021  & 0.46  & 1.97 & 2.16 & Yes \\ \hline
Ours  & \textbf{0.52} &  \textbf{1.32} & \textbf{1.54} & Yes   \\
\bottomrule
\end{tabular}
\end{table}

More results are in the appendix: 
Section \ref{sec:diffskin} analyzes (predicted) oxygen value distributions in different race groups;
Section \ref{sec:extra-results} has more visualization on our model's performance on different datasets, patients with different relevant diseases.

\section{Concluding Remarks} \label{sec:conclusion}
This paper introduces the new task of inferring oxygen saturation from radio signals. It develops a new gated transformer architecture to deliver this application and adapt deep models to auxiliary categorical variables. We note the work have some limitations. First, the paper focuses on special use cases (e.g., at-home oxygen monitoring in very old adults or during sleep), but is not suitable for other use cases (e.g., measuring oxygen to optimize performance during exercise).  Second, the results in the paper provide an initial proof of concept that the shape and dynamics of the breathing signal include sufficient clues to infer a useful estimate of oxygen saturation. However, before this system can be used in clinical care, one needs clinical studies to quantify the performance for different disease conditions.

\bibliographystyle{plain}
\bibliography{neurips_ts4h_2022}

\begin{thebibliography}{10}

\bibitem{adib2015smart}
Fadel Adib, Hongzi Mao, Zachary Kabelac, Dina Katabi, and Robert~C Miller.
\newblock Smart homes that monitor breathing and heart rate.
\newblock In {\em Proceedings of the 33rd annual ACM conference on human
  factors in computing systems}, 2015.

\bibitem{al2019remote}
Mohammed Al-Khafajiy, Thar Baker, Carl Chalmers, Muhammad Asim, Hoshang
  Kolivand, Muhammad Fahim, and Atif Waraich.
\newblock Remote health monitoring of elderly through wearable sensors.
\newblock {\em Multimedia Tools and Applications}, 78(17):24681--24706, 2019.

\bibitem{bal2015non}
Ufuk Bal.
\newblock Non-contact estimation of heart rate and oxygen saturation using
  ambient light.
\newblock {\em Biomedical optics express}, 6(1):86--97, 2015.

\bibitem{na1961}
N.A. Bergman.
\newblock Cyclic variations in blood oxygenation with the respiratory cycle.
\newblock {\em Anesthesiology}, 1961.

\bibitem{MROS}
Terri Blackwell, Kristine Yaffe, Sonia Ancoli-Israel, Susan Redline, Kristine~E
  Ensrud, Marcia~L Stefanick, Alison Laffan, Katie~L Stone, and
  Osteoporotic~Fractures in~Men Study~Group.
\newblock Associations between sleep architecture and sleep-disordered
  breathing and cognition in older community-dwelling men: the osteoporotic
  fractures in men sleep study.
\newblock {\em Journal of the American Geriatrics Society}, 59(12):2217--2225,
  2011.

\bibitem{celler1995remote}
BG~Celler, W~Earnshaw, ED~Ilsar, L~Betbeder-Matibet, MF~Harris, R~Clark,
  T~Hesketh, and NH~Lovell.
\newblock Remote monitoring of health status of the elderly at home. a
  multidisciplinary project on aging at the university of new south wales.
\newblock {\em International journal of bio-medical computing}, 40(2):147--155,
  1995.

\bibitem{MESA}
Xiaoli Chen, Rui Wang, Phyllis Zee, Pamela~L Lutsey, Sogol Javaheri, Carmela
  Alc{\'a}ntara, Chandra~L Jackson, Michelle~A Williams, and Susan Redline.
\newblock Racial/ethnic differences in sleep disturbances: the multi-ethnic
  study of atherosclerosis (mesa).
\newblock {\em Sleep}, 38(6), 2015.

\bibitem{choi2016severity}
Eunkyung Choi, Doo-Heum Park, Jae-hak Yu, Seung-Ho Ryu, and Ji-Hyeon Ha.
\newblock The severity of sleep disordered breathing induces different decrease
  in the oxygen saturation during rapid eye movement and non-rapid eye movement
  sleep.
\newblock {\em Psychiatry investigation}, 13(6):652, 2016.

\bibitem{bert}
Jacob Devlin, Ming-Wei Chang, Kenton Lee, and Kristina Toutanova.
\newblock Bert: Pre-training of deep bidirectional transformers for language
  understanding.
\newblock {\em arXiv preprint arXiv:1810.04805}, 2018.

\bibitem{diaz2002usefulness}
Genaro D{\'\i}az-Rega{\~n}{\'o}n, Eduardo Mi{\~n}ambres, Marisol Holanda,
  Segundo Gonz{\'a}lez-Herrera, Francisco L{\'o}pez-Espadas, and Carlos
  Garrido-D{\'\i}az.
\newblock Usefulness of venous oxygen saturation in the jugular bulb for the
  diagnosis of brain death: report of 118 patients.
\newblock {\em Intensive care medicine}, 28(12):1724--1728, 2002.

\bibitem{ding2018measuring}
Xinyi Ding, Damoun Nassehi, and Eric~C Larson.
\newblock Measuring oxygen saturation with smartphone cameras using
  convolutional neural networks.
\newblock {\em IEEE journal of biomedical and health informatics},
  23(6):2603--2610, 2018.

\bibitem{fan2020learning}
Lijie Fan, Tianhong Li, Rongyao Fang, Rumen Hristov, Yuan Yuan, and Dina
  Katabi.
\newblock Learning longterm representations for person re-identification using
  radio signals.
\newblock In {\em Proceedings of the IEEE/CVF Conference on Computer Vision and
  Pattern Recognition}, 2020.

\bibitem{fan2020home}
Lijie Fan, Tianhong Li, Yuan Yuan, and Dina Katabi.
\newblock In-home daily-life captioning using radio signals.
\newblock In {\em ECCV}, pages 105--123. Springer, 2020.

\bibitem{feiner2007dark}
John~R Feiner, John~W Severinghaus, and Philip~E Bickler.
\newblock Dark skin decreases the accuracy of pulse oximeters at low oxygen
  saturation: the effects of oximeter probe type and gender.
\newblock {\em Anesthesia \& Analgesia}, 2007.

\bibitem{gries1996normal}
Robert~E Gries and Lee~J Brooks.
\newblock Normal oxyhemoglobin saturation during sleep: how low does it go?
\newblock {\em Chest}, 110(6):1489--1492, 1996.

\bibitem{guazzi2015non}
Alessandro~R Guazzi, Mauricio Villarroel, Joao Jorge, Jonathan Daly, Matthew~C
  Frise, Peter~A Robbins, and Lionel Tarassenko.
\newblock Non-contact measurement of oxygen saturation with an rgb camera.
\newblock {\em Biomedical optics express}, 6(9):3320--3338, 2015.

\bibitem{han2020}
Dong Han, Chenyang Wang, Xiaojing Feng, and Jing Wu.
\newblock Delirium during recovery in patients with severe covid-19: Two case
  reports.
\newblock {\em Frontiers in Medicine}, 7, 2020.

\bibitem{hanson1975}
PG~Hanson, KH~Lin, and MB~McIlroy.
\newblock Influence of breathing pattern on oxygen exchange during hypoxia and
  exercise.
\newblock In {\em J Appl Physiol}, June 1975.

\bibitem{wigait}
Chen-Yu Hsu, Yuchen Liu, Zachary Kabelac, Rumen Hristov, Dina Katabi, and
  Christine Liu.
\newblock Extracting gait velocity and stride length from surrounding radio
  signals.
\newblock In {\em Proc. of the 2017 CHI Conf. on Human Factors in Computing
  Systems}, 2017.

\bibitem{kiyasseh2020clops}
Dani Kiyasseh, Tingting Zhu, and David~A Clifton.
\newblock Clops: Continual learning of physiological signals.
\newblock {\em arXiv preprint arXiv:2004.09578}, 2020.

\bibitem{lapinsky1999safety}
SE~Lapinsky, M~Aubin, S~Mehta, P~Boiteau, and AS~Slutsky.
\newblock Safety and efficacy of a sustained inflation for alveolar recruitment
  in adults with respiratory failure.
\newblock {\em Intensive care medicine}, 25(11):1297--1301, 1999.

\bibitem{lecube2009diabetes}
Albert Lecube, Gabriel Sampol, Patricia Lloberes, Odile Romero, Jordi Mesa,
  Cristina Hern{\'a}ndez, and Rafael Sim{\'o}.
\newblock Diabetes is an independent risk factor for severe nocturnal hypoxemia
  in obese patients. a case-control study.
\newblock {\em PloS one}, 4(3):e4692, 2009.

\bibitem{li2022targeted}
Tianhong Li, Peng Cao, Yuan Yuan, Lijie Fan, Yuzhe Yang, Rogerio~S Feris, Piotr
  Indyk, and Dina Katabi.
\newblock Targeted supervised contrastive learning for long-tailed recognition.
\newblock In {\em Proceedings of the IEEE/CVF Conference on Computer Vision and
  Pattern Recognition}, pages 6918--6928, 2022.

\bibitem{li2020addressing}
Tianhong Li, Lijie Fan, Yuan Yuan, Hao He, Yonglong Tian, Rogerio Feris, Piotr
  Indyk, and Dina Katabi.
\newblock Addressing feature suppression in unsupervised visual
  representations.
\newblock {\em arXiv e-prints}, pages arXiv--2012, 2020.

\bibitem{Li_2022_WACV}
Tianhong Li, Lijie Fan, Yuan Yuan, and Dina Katabi.
\newblock Unsupervised learning for human sensing using radio signals.
\newblock In {\em Proceedings of the IEEE/CVF Winter Conference on Applications
  of Computer Vision (WACV)}, January 2022.

\bibitem{liu2019auxiliary}
Yifan Liu, Bohan Zhuang, Chunhua Shen, Hao Chen, and Wei Yin.
\newblock Auxiliary learning for deep multi-task learning.
\newblock {\em arXiv preprint arXiv:1909.02214}, 2019.

\bibitem{mathew2021remote}
Joshua Mathew, Xin Tian, Min Wu, and Chau-Wai Wong.
\newblock Remote blood oxygen estimation from videos using neural networks.
\newblock {\em arXiv preprint arXiv:2107.05087}, 2021.

\bibitem{mordan2018revisiting}
Taylor Mordan, Nicolas Thome, Gilles Henaff, and Matthieu Cord.
\newblock Revisiting multi-task learning with rock: a deep residual auxiliary
  block for visual detection.
\newblock In {\em NeurIPS}, 2018.

\bibitem{moss2005compromised}
Mark Moss, Mark Franks, Pamela Briggs, David Kennedy, and Andrew Scholey.
\newblock Compromised arterial oxygen saturation in elderly asthma sufferers
  results in selective cognitive impairment.
\newblock {\em Journal of Clinical and Experimental Neuropsychology},
  27(2):139--150, 2005.

\bibitem{murugappan2010classification}
Murugappan Murugappan, Nagarajan Ramachandran, Yaacob Sazali, et~al.
\newblock Classification of human emotion from eeg using discrete wavelet
  transform.
\newblock {\em Journal of biomedical science and engineering}, 3(04):390, 2010.

\bibitem{narayan2017neural}
Shashi Narayan, Nikos Papasarantopoulos, Shay~B Cohen, and Mirella Lapata.
\newblock Neural extractive summarization with side information.
\newblock {\em arXiv preprint arXiv:1704.04530}, 2017.

\bibitem{nlm}
NLM.
\newblock Aging changes in the lungs.
\newblock \url{https://medlineplus.gov/ency/article/004011.htm}, 2020.
\newblock Accessed: 2020-10-01.

\bibitem{palma2008oxygen}
David~T Palma, George~M Philips, Miguel~R Arguedas, Susan~M Harding, and
  Michael~B Fallon.
\newblock Oxygen desaturation during sleep in hepatopulmonary syndrome.
\newblock {\em Hepatology}, 47(4):1257--1263, 2008.

\bibitem{Parkins1998}
K.~J. Parkins, L.~M. Poets, C. F.and~O'Brien, V.~A. Stebbens, and D.~P.
  Southall.
\newblock Effect of exposure to 15\% oxygen on breathing patterns and oxygen
  saturation in infants: interventional study.
\newblock {\em BMJ (Clinical research ed.)}, 1998.

\bibitem{rahman2015dopplesleep}
Tauhidur Rahman, Alexander~T Adams, Ruth~Vinisha Ravichandran, Mi~Zhang,
  Shwetak~N Patel, Julie~A Kientz, and Tanzeem Choudhury.
\newblock Dopplesleep: A contactless unobtrusive sleep sensing system using
  short-range doppler radar.
\newblock In {\em UbiComp}, pages 39--50, 2015.

\bibitem{reybrouck1999gender}
Tony Reybrouck and Robert Fagard.
\newblock Gender differences in the oxygen transport system during maximal
  exercise in hypertensive subjects.
\newblock {\em Chest}, 115(3):788--792, 1999.

\bibitem{rim2020deep}
Beanbonyka Rim, Nak-Jun Sung, Sedong Min, and Min Hong.
\newblock Deep learning in physiological signal data: A survey.
\newblock {\em Sensors}, 20(4):969, 2020.

\bibitem{unet}
Olaf Ronneberger, Philipp Fischer, and Thomas Brox.
\newblock U-net: Convolutional networks for biomedical image segmentation.
\newblock In {\em International Conference on Medical image computing and
  computer-assisted intervention}, 2015.

\bibitem{shao2015noncontact}
Dangdang Shao, Chenbin Liu, Francis Tsow, Yuting Yang, Zijian Du, Rafael Iriya,
  Hui Yu, and Nongjian Tao.
\newblock Noncontact monitoring of blood oxygen saturation using camera and
  dual-wavelength imaging system.
\newblock {\em IEEE Transactions on Biomedical Engineering}, 2015.

\bibitem{shen2016automatic}
Xiaoyong Shen, Aaron Hertzmann, Jiaya Jia, Sylvain Paris, Brian Price, Eli
  Shechtman, and Ian Sachs.
\newblock Automatic portrait segmentation for image stylization.
\newblock In {\em Computer Graphics Forum}, volume~35, pages 93--102, 2016.

\bibitem{sjoding2020racial}
Michael~W Sjoding, Robert~P Dickson, Theodore~J Iwashyna, Steven~E Gay, and
  Thomas~S Valley.
\newblock Racial bias in pulse oximetry measurement.
\newblock {\em New England Journal of Medicine}, 383(25):2477--2478, 2020.

\bibitem{rf-fall}
Yonglong Tian, Guang-He Lee, Hao He, Chen-Yu Hsu, and Dina Katabi.
\newblock Rf-based fall monitoring using convolutional neural networks.
\newblock {\em Proceedings of the ACM on Interactive, Mobile, Wearable and
  Ubiquitous Technologies}, 2(3):1--24, 2018.

\bibitem{van2016new}
Mark Van~Gastel, Sander Stuijk, and Gerard De~Haan.
\newblock New principle for measuring arterial blood oxygenation, enabling
  motion-robust remote monitoring.
\newblock {\em Scientific reports}, 6(1):1--16, 2016.

\bibitem{van2019data}
Mark van Gastel, Wim Verkruysse, and Gerard de~Haan.
\newblock Data-driven calibration estimation for robust remote pulse-oximetry.
\newblock {\em Applied Sciences}, 9(18):3857, 2019.

\bibitem{wang2022self}
Yangtao Wang, Xi~Shen, Shell~Xu Hu, Yuan Yuan, James~L Crowley, and Dominique
  Vaufreydaz.
\newblock Self-supervised transformers for unsupervised object discovery using
  normalized cut.
\newblock In {\em Proceedings of the IEEE/CVF Conference on Computer Vision and
  Pattern Recognition}, pages 14543--14553, 2022.

\bibitem{wang2022tokencut}
Yangtao Wang, Xi~Shen, Yuan Yuan, Yuming Du, Maomao Li, Shell~Xu Hu, James~L
  Crowley, and Dominique Vaufreydaz.
\newblock Tokencut: Segmenting objects in images and videos with
  self-supervised transformer and normalized cut.
\newblock {\em arXiv preprint arXiv:2209.00383}, 2022.

\bibitem{yang2022artificial}
Yuzhe Yang, Yuan Yuan, Guo Zhang, Hao Wang, Ying-Cong Chen, Yingcheng Liu,
  Christopher~G Tarolli, Daniel Crepeau, Jan Bukartyk, Mithri~R Junna, et~al.
\newblock Artificial intelligence-enabled detection and assessment of
  parkinson’s disease using nocturnal breathing signals.
\newblock {\em Nature medicine}, pages 1--9, 2022.

\bibitem{yuan2017temporal}
Yuan Yuan, Xiaodan Liang, Xiaolong Wang, Dit-Yan Yeung, and Abhinav Gupta.
\newblock Temporal dynamic graph lstm for action-driven video object detection.
\newblock In {\em Proceedings of the IEEE international conference on computer
  vision}, pages 1801--1810, 2017.

\bibitem{yuan2019marginalized}
Yuan Yuan, Yueming Lyu, Xi~Shen, Ivor~W Tsang, and Dit-Yan Yeung.
\newblock Marginalized average attentional network for weakly-supervised
  learning.
\newblock {\em arXiv preprint arXiv:1905.08586}, 2019.

\bibitem{yue2018extracting}
Shichao Yue, Hao He, Hao Wang, Hariharan Rahul, and Dina Katabi.
\newblock Extracting multi-person respiration from entangled rf signals.
\newblock In {\em UbiComp}, 2018.

\bibitem{SHHS2}
Guo-Qiang Zhang, Licong Cui, Remo Mueller, Shiqiang Tao, Matthew Kim, Michael
  Rueschman, Sara Mariani, Daniel Mobley, and Susan Redline.
\newblock The national sleep research resource: towards a sleep data commons.
\newblock {\em Journal of the American Medical Informatics Association}, 2018.

\bibitem{eq-radio}
Mingmin Zhao, Fadel Adib, and Dina Katabi.
\newblock Emotion recognition using wireless signals.
\newblock In {\em MobiCom}, 2016.

\bibitem{zhao2017learning}
Mingmin Zhao, Shichao Yue, Dina Katabi, Tommi~S Jaakkola, and Matt~T Bianchi.
\newblock Learning sleep stages from radio signals: A conditional adversarial
  architecture.
\newblock In {\em ICML}, 2017.

\end{thebibliography}

\newpage
\appendix

\newpage
\section{Related Work}
\paragraph{Monitoring Oxygen Saturation.}
The most accurate measurements of oxygen saturation are invasive and require arterial blood samples. The non-invasive and widely-used method for measuring  oxygen saturation~(SpO2) uses a pulse oximeter, a small device worn on the finger. 
To enable remote SpO2 measurements, past work has investigated the use of cameras~\cite{van2016new,van2019data,shao2015noncontact,bal2015non,guazzi2015non}. However those methods have limitations like susceptibility to noise, sensitivity to motion, and a need for ambient light. 
Recently, deep learning has been considered to aid SpO2 monitoring using cameras.  \cite{ding2018measuring} tried to monitor SpO2 using smartphones. But their solution requires the fingertip to be pressed against the camera, and hence cannot provide continuous overnight measurements. A more recent work~\cite{mathew2021remote} has estimated SpO2 in a contactless way with regular RGB cameras. Their method first extracts the region of interest from the video of the person's palm, then uses a CNN model to estimate SpO2. While this approach is contactless, it still requires the user to keep their hand in front of the video camera for the duration of the monitoring, which is not practical for continuous or overnight monitoring. Our work differs from all of these prior works in that we predict oxygen values from breathing or radio waves, which allows for continuous oxygen sensing in a contactless and passive manner.


\paragraph{Contactless Health Sensing with Radio Signals.} 
The past decade has seen a rapid growth in research on passive sensing using RF signals. Early work has demonstrated the possibility of sensing one's breathing and heart rate using radio signals~\cite{adib2015smart}. Later, researchers have shown that by analyzing the RF signals that bounce off the human body, they can monitor a variety of health metrics including sleep, gait, falls, and even human emotions~\cite{zhao2017learning, wigait,rf-fall,eq-radio,fan2020learning,Li_2022_WACV,yang2022artificial,li2020addressing}. We build on this work to enable SpO2 monitoring with radio waves. 


\paragraph{Adaptation of ML Models to Medical Indices.}
Prior deep learning models~\cite{wang2022tokencut,wang2022self,li2022targeted,yuan2019marginalized,yuan2017temporal} do not adapt to a person's medical indices. For example, the literature has models that infer sleep stages from respiration~\cite{zhao2017learning}, detect arrhythmia from ECG~\cite{kiyasseh2020clops}, and classify emotion from EEG signals~\cite{murugappan2010classification}. A recent survey~\cite{rim2020deep} collected 147 papers about learning with physiological signals. None of the deep models therein adapt to a person's medical indices. 
The deep learning literature includes a few approaches for leveraging auxiliary variables. If the variable is available at inference time, typically it is taken as an extra input~\cite{narayan2017neural,shen2016automatic}. Variables accessible only during training are typically used as extra supervisors to regularize the model via multi-task learning~\cite{liu2019auxiliary,mordan2018revisiting}. 
We propose a gating mechanism to handle categorical variables and show that it performs better.
\section{Extracting Breathing Signals from RF Signals}\label{sup:br-from-rf}
We leverage past work on extracting breathing from the RF signals. Specifically, our system is equipped with a multi-antenna Frequency-Modulated Continuous-Wave (FMCW) radio, which is commonly used in passive health monitoring~\cite{rahman2015dopplesleep, yue2018extracting,fan2020home}. The radio transmits a very low power RF signal and captures its reflections from the environment. We process these reflections using the algorithm in~\cite{yue2018extracting} to infer the subject's breathing signal. Past work shows that breathing signals extracted in this manner are highly accurate. Specifically, their correlation with an FDA-approved breathing belt on the person ranges from $91\%$ to $99\%$, depending on the distance from the radio and the distance between people~\cite{yue2018extracting}. 

\section{Backbone Model: BERT-UNet}\label{model:basemodel}
Our backbone model a combination of a BERT module~\cite{bert} and UNet~\cite{unet}. As shown in \Figref{fig:model}(a), our \emph{BERT-UNet} model consists of an encoder $E(\cdot;\theta_e)$ and an oxygen predictor $F(\cdot;\theta_f)$. The encoder is composed of a fully convolutional network (FCN) followed by a bidirectional-transformer (BERT) module~\cite{bert}. The FCN extracts local features from the raw respiration signals, then the BERT module captures long-term temporal dependencies based on those features. 
The predictor $F$ is composed of several deconvolutional layers, which up-sample the extracted features to the same time resolution of oxygen saturation. 
Formally, we have $E: \sR^{1\times f_{b}T} \to \sR^{n \times \alpha f_{b}T}$ and $F:  \sR^{n \times \alpha f_{b}T} \to \sR^{1\times f_{o}T}$ where $n$ is the dimension of the respiration feature and $\alpha$ is the down-sampling factor ($\alpha = 1/240$ in our experiments). 

The model is trained with a combination of the $L_1$ loss and the correlation loss given below:


\vspace{-2mm}
\begin{equation}
\resizebox{0.45\hsize}{!}{$
    \gL(\hat{y}, y) = \frac{\|\hat{y} - y \|_1}{f_o T} -
    \lambda \frac{\sum_{i} (\hat{y}^i - \mu_{\hat{y}})(y^i - \mu_{y})}{\sqrt{\sum_{i}(\hat{y}^i - \mu_{\hat{y}})^2\sum_{i}(y^i - \mu_{y})^2}}.
$}
\label{eq:l1-loss}
\end{equation}
Here $\hat{y}=F(E(x; \theta_e);\theta_f)$ is the model prediction, $y$ is the ground truth oxygen, $\mu_{y}$ and $\mu_{\hat{y}}$ are the mean values of $y$ and $\hat{y}$, and $\lambda$ is a hyper-parameter to balance the two loss terms. We choose the $L_1$ loss over other regression loss functions, since it is more robust to outliers and empirically has better performance. We also use the correlation loss to help in matching the fluctuations in the predicted oxygen with the fluctuations of the ground truth.

\begin{figure}[h!]
\centering
\includegraphics[width=0.9\columnwidth]{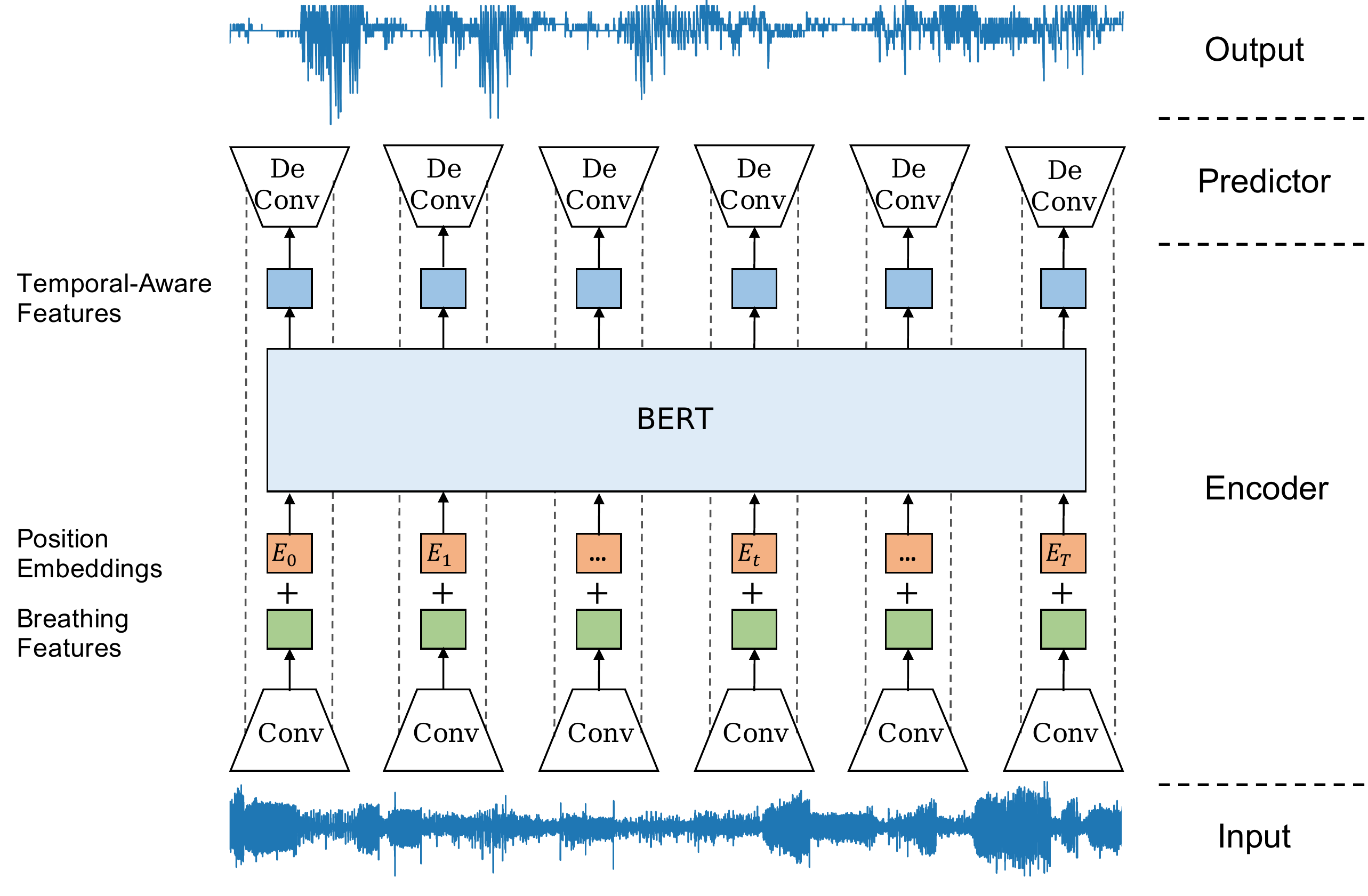}
\caption{Network architecture for the backbone BERT-UNet model.}
\label{fig:vanilla}
\end{figure}

\paragraph{Architecture Specifications.}\label{sup:implement}
The encoder has nine 1-D convolutional layers (Conv-BatchNorm-RReLU) that shrink the features' temporal dimension by 240 times. It is then followed by several bi-directional multi-head self-attention layers (BERT)~\cite{bert} to aggregate the temporal information at the bottleneck. We use 8 layers, 6 heads with hidden-size of 256, intermediate-size of 512 for self-attention, and the max position embeddings is 2400. The decoder contains 7 layers of 1-D de-convolutional layers (DeConv-Norm-RReLU). We also use a skip connection~\cite{unet} by concatenating the convolutional layers in the encoder to the de-convolutional layers in the predictor. Figure~\ref{fig:vanilla} illustrates the overall network architecture. 

\section{Full Model: Gated BERT-UNet}\label{sec:fullmodel}

In medical applications, there are useful side variables. Adapting to such variables will likely improve performance and make the results more personalized. In many cases, the relevant variables are binary or categorical. For example, the relevant variables for oxygen saturation include gender, whether the person is a smoker, whether they have asthma, etc. The categorical nature of these variables induces \textbf{discontinuity} in the learned function over the physiological indices. Specifically, consider oxygen saturation $y=g_s(x)$ as a function of respiration $x$ and gender $s$ ($0$ for male and $1$ for female). $g_0(x)$ and $g_1(x)$ can be two different functions since men and women have differences in their oxygen transport systems~\cite{reybrouck1999gender}. 

To handle such discontinuity for better leveraging the categorical variables, we propose \emph{Gated BERT-UNet}, a new model that augments BERT-UNet with multiple predictive heads. It selects the most suitable head for a person via a gate controlled by the subject's categorical indices. The model supports both variables available at the time of inference (e.g., gender), as well as dense categorical variables concurrently learned from the input signals (e.g., sleep stages).
 
\Figref{fig:model}(b) illustrates the model. It has a gate function $G(v,u): \gV \times \gU \to \{1,2,\cdots,N\}$ where $v \in \gV$ and $u \in \gU$ are accessible/inaccessible variables, $N$ is the number of gate statuses. We use the term \emph{accessible variable} for variables easily available during inference time, e.g., gender, and the term \emph{inaccessible variable} for information that is not available during inference, but available during training, like a person's sleep stages. 
Inaccessible variables are typically dense time series (e.g., sleep stages). Their prediction are learned concurrently with the main task under  full supervision. 
The construction of the gate function $G(v,u)$ is described in the next sub-section.

The model has $N$ heads $\{F_i\}_{i=1}^N$ that adapt the prediction $\hat{y}_i=F_i(E(x))$ to the gate status. It also has an extra predictor $F_u$ to infer inaccessible variables $u$. During testing time, based on the accessible variables $v$ and estimated inaccessible variables $\hat{u}$, we evaluate the gate status $s = G(v,\hat{u})$. 

In the case of oxygen prediction, $\hat{y}_i$ and the gate status $s$ are time series. As shown in \Figref{fig:model}(b), the final prediction at each time step, is the gated combination of every head's output, i.e.
$ \forall t=1,\dots, f_oT, \hat{y}^t=\sum_{i=1}^{N}\1[s^t=i]\hat{y}^t_{i}$ .
We train Gated BERT-UNet (GBU) with the following loss,
\vspace{-1mm}
\begin{equation}
\gL_{\texttt{GBU}}(\hat{y},\hat{u}, y, u) = \gL(\hat{y}, y) +  \frac{\lambda_u}{f_oT}\sum_{t=1}^{f_oT}\gL_{\texttt{CE}}(\hat{u}^t, u^t),
\end{equation}
where $\gL$ is the main loss defined in \Eqref{eq:l1-loss}, $\gL_{\texttt{CE}}$ is the cross-entropy loss to train the branch for predicting inaccessible variables and $\lambda_u$ is a balancing hyperparameter.

\paragraph{Mapping Variables to Heads.}\label{sec:grad-sim}
The number of heads in a Gated BERT-UNet model puts an upper bound on the number of possible gate states. For example, if the model has 6 heads, the gate can take only 1 of 6 states. Typically, we have many more variable states than gate states. To find a proper mapping $G(v,u)$ from variable state to gate state, we rely on gradient similarity. For example, if we want to check whether male smokers should be in the same group as female smokers, we take a pretrained backbone BERT-UNet and compute its averaged gradient (w.r.t the loss function) over all male smokers and all female smokers in the dataset. Then we check the cosine similarity between the two gradients. If the gradients are similar, which means the two categories move the loss function in the same direction, we can use the same predictor for them. On the other hand, if the gradients are vastly different, it is preferable to separate such categories and assign them to different gate statuses.

\section{Datasets and Metrics} \label{sec:datasets}

\paragraph{Medical Datasets.}
We leverage three public medical datasets: Sleep Heart Health Study (\textit{SHHS})~\cite{SHHS2}, Multi-Ethnic Study of Atherosclerosis (\textit{MESA})~\cite{MESA}, and Osteoporotic Fractures in Men Study (\textit{MrOS})~\cite{MROS}. 
The datasets were collected during sleep studies. For each subject, they include the respiration signals throughout the night along with the corresponding blood oxygen time series.
The breathing signals are collected using a breathing belt around the chest or abdomen, and the oxygen is measured using a pulse oximeter.
The datasets also contain side variables including sleep stages, which for every time instance assign to the subject one of the following: Awake, Rapid Eye Movement (REM) or non-REM stage. 

We note that the subjects in these studies have an age range between 40 and 95, and some of them suffer from a variety of diseases such as chronic bronchitis, cardiovascular diseases, and diabetes. This allows for a wider range of oxygen variability beyond the typical range of healthy individuals.    

\begin{wrapfigure}{r}{0.3\textwidth}
\centering
\vspace{-4mm}
\includegraphics[width=0.3\columnwidth]{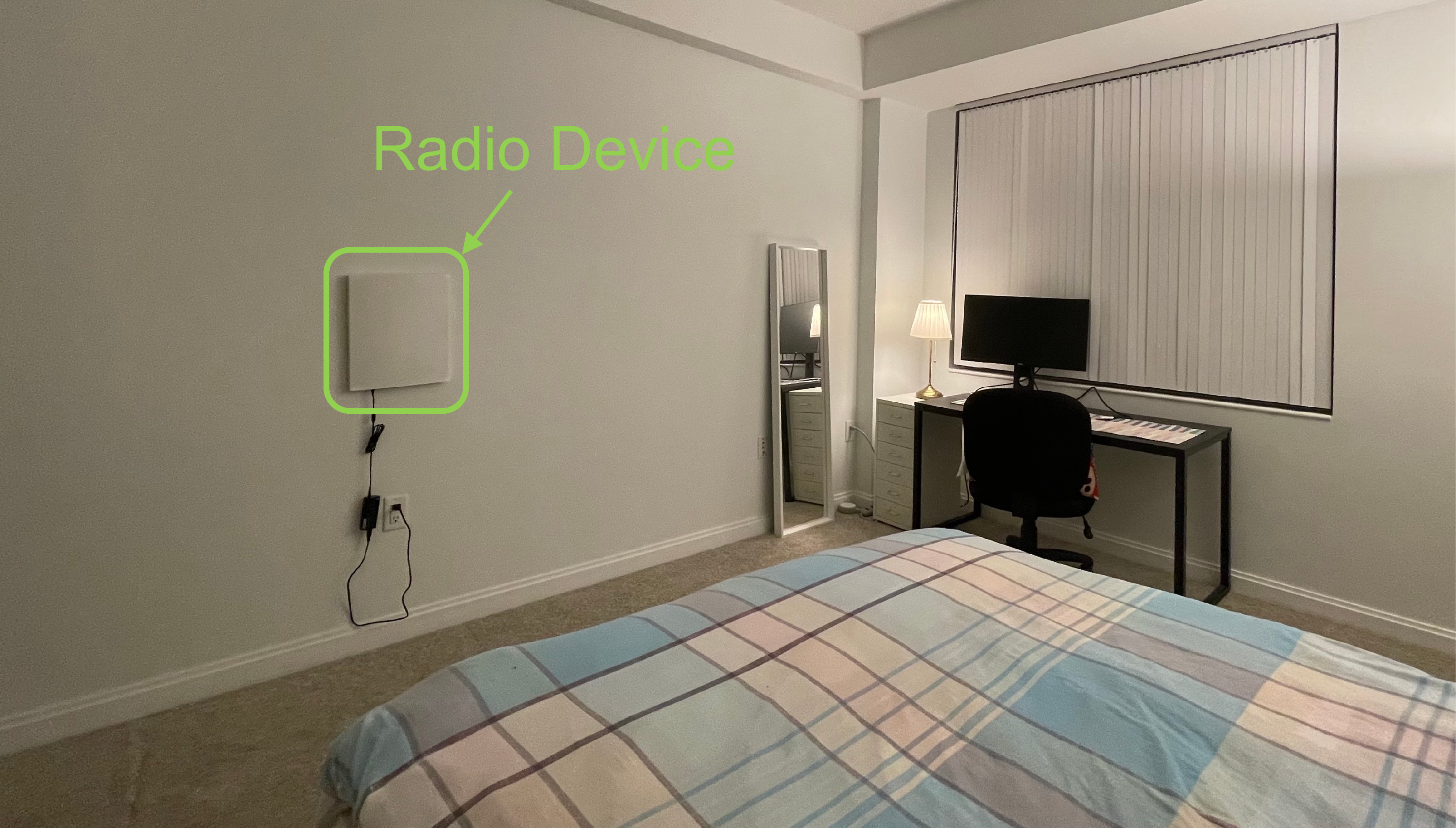}
\vspace{-4mm}
\caption{The radio device to collect RF signals.}
\vspace{-6mm}
\label{fig:radio}
\end{wrapfigure}

\paragraph{RF Datasets.} We collected a dataset of RF signals paired with SpO2 measurements. The data is collected from two sleep labs. In total, there are 400 hours of data from 49 overnight recordings of 32 subjects. The dataset contains subjects of different genders and races. Some subjects are healthy volunteers while others are patients with sleep problems. Thus, the ground truth oxygen saturation distribution is wider than normal ranges. A radio device is installed in the room to collect RF signals, as shown in  Fig.~\ref{fig:radio}. The radio signals are synchronized with the SpO2 measurements and processed with the algorithm in~\cite{yue2018extracting} to extract the person's breathing signals.

\paragraph{Metrics.}
Let $\hat{y}$, $y$ denote the predicted and ground truth oxygen saturation. Following the previous work~\cite{mathew2021remote}, we use three standard metrics for evaluation: (1)~Correlation: \resizebox{0.2\hsize}{!}{$\frac{\sum_{t} (y^t-\mu_y)(\hat{y}^t-\mu_{\hat{y}}) }{\sqrt{\sum_{t} (y^t-\mu_y)^2 \sum_t(\hat{y}^t-\mu_{\hat{y}})^2}}$}, where $\mu_y$ and $\mu_{\hat{y}}$ are the averaged oxygen saturations; (2)~MAE: \resizebox{0.15\hsize}{!}{$\frac{1}{T} \sum_{t=1}^T |y^t - \hat{y}^t|$}; and (3)~RMSE: \resizebox{0.15\hsize}{!}{$\sqrt{\frac{1}{T} \sum_{t=1}^T (y^t - \hat{y}^t)^2}$}. 
The models take a night of breathing as input and predict the corresponding oxygen levels. During the evaluation, we divide the model's prediction and ground truth oxygen into 240-second segments to compute the metrics. We note that those metrics are sensitive to the segment's length. For a fair comparison, we follow \cite{mathew2021remote} and use 240-second intervals.

\section{Baselines.}\label{sec:baseline}
We compare the following neural network architectures for our backbone model: (a)~\textit{CNN} is a fully convolutional model composed of eight 1-D convolutional layers and seven 1-D deconvolutional layers; (b)~\textit{CNN-RNN} augments the CNN model with a recurrent unit (one layer LSTM) in the bottleneck to better captures the long-term temporal relationships of the data. (c)~\textit{BERT-UNet} further makes two improvements on the CNN-RNN model. First, it uses an attention module to replace recurrent unit for temporal modelling. Second, it adds skip links between encoding convolutional layers and decoding deconvolutional layers at the same temporal scale to better capture the signal's local information. 

To evaluate our design for incorporating side information, we compare the following models:
(a) \textit{BERT-UNet + VarAug}, which uses BERT-UNet as backbone and takes accessible variables as extra inputs and inaccessible variables as auxiliary tasks. (b) Our \textit{Gated BERT-UNet} model, which uses a multi-head model gated by physiological variables.

\section{Training and Evaluation Protocols}\label{sec:protocol}

\paragraph{Train/Valid/Test Spilt.} Due to the limited data amount in the RF dataset, the RF data is all hold out for testing. In our study, all models are trained on the medical datasets and evaluated on both the medical datasets and the RF dataset. Collectively, the medical datasets have about 48,000 hours of data from 5,765 subjects in total.  We randomly split subjects, 70\% for training and validation, and 30\% for testing, and fix the splits in all experiments. We train each model on the union of the training sets from the three medical datasets, and tested on each test set. 

\paragraph{Side Variable Specifications.} For models that incorporate side variables, we use \emph{gender} as the accessible variable and \emph{sleep stages} as the inaccessible variable. In Gated BERT-UNet, we use gradient similarity to map gate status as described in the method section, which results in the following $6$ categories: (male, awake), (male, REM sleep), (male, 
non-REM sleep), (female, awake), (female, REM sleep), (female, non-REM sleep). The sleep stages themselves are learned from the input since they are an inaccessible variable. In the baseline VarAug, the gender variable is provided as an additional input and the sleep stages are used as an auxiliary task in a multitask model. 

\paragraph{Implementation Details.}
We implement all models in PyTorch. All experiments are carried out on a NVIDIA TITAN Xp GPU with 12 GB memory. The number of parameters for the Gated BERT-UNet model is 26,821,113 and the model size is 107.28MB. In the training process, we use the Adam optimizer with a learning rate of $2 \times 10^{-4}$, and train the model for 500 epochs. Due to the varying input length, we set the batch size for all models to 1 (i.e., one night of respiration signal and the corresponding oxygen time series). 

\section{Results for Different Skin Colors}\label{sec:diffskin}

\begin{figure}[t]
\begin{center}
\includegraphics[width=0.5\columnwidth]{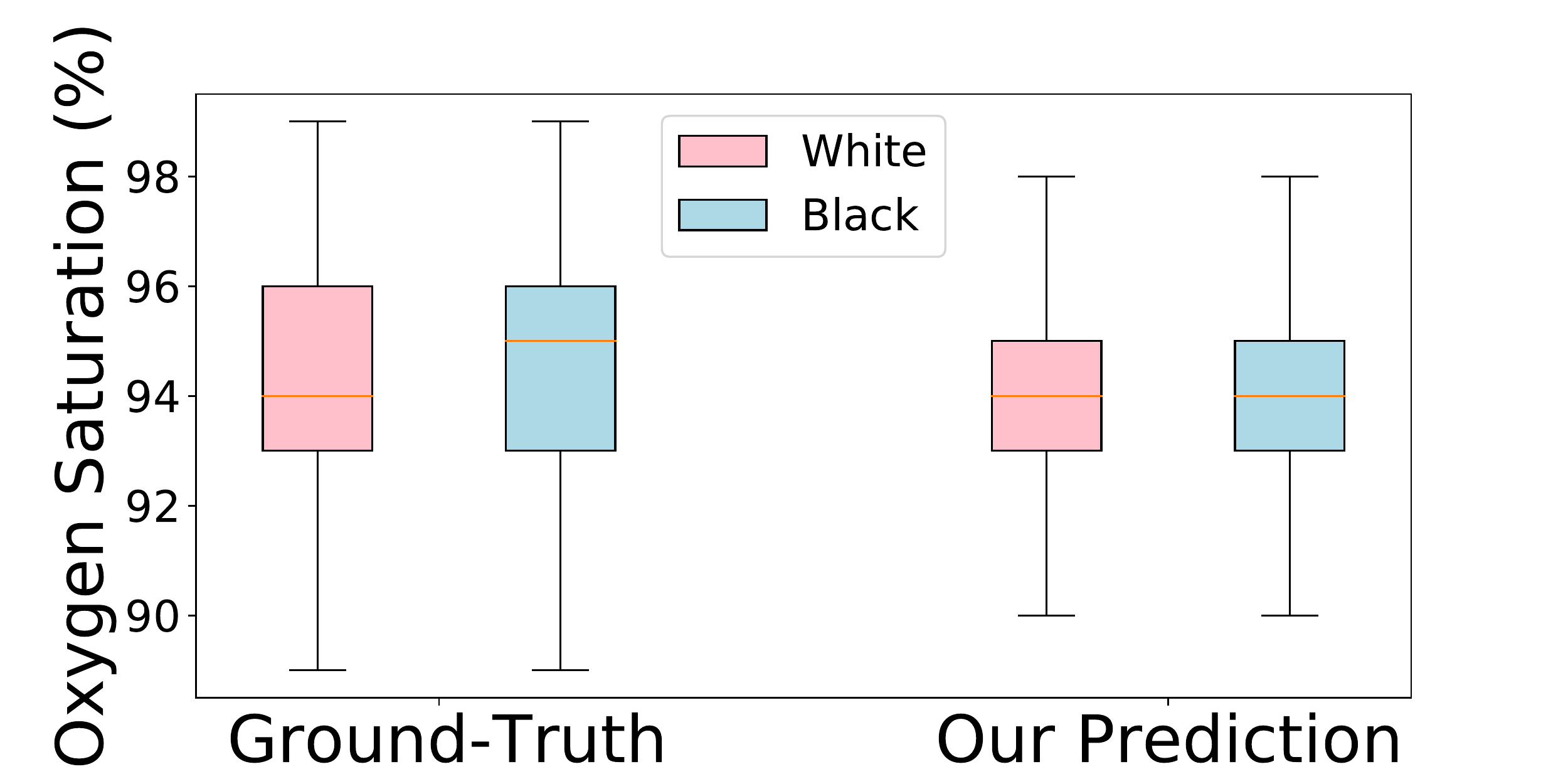}
\end{center}
\caption{Box plot compares the ground-truth oxygen saturation measured by pulse oximeter and our predicted oxygen saturation for two races.} 
\label{fig:race-result}
\end{figure}

Since pulse oximeters rely on measuring light absorbance through the finger, they are known to be affected by skin color and tend to overestimate blood oxygen saturation in subjects with dark skin~\cite{feiner2007dark,sjoding2020racial}. A large study that looked at tens of thousands of white and black COVID patients found that the ``reliance on pulse oximetry to triage patients and adjust supplemental oxygen levels may place black patients at increased risk for hypoxaemia"~\cite{sjoding2020racial}.  In contrast, breathing and RF signals have no intrinsic bias against skin color.  \Figref{fig:race-result} shows the distributions of the oximetry-based ground-truth oxygen and the Gated BERT-UNet prediction for different races, for the union of all datasets. The ground-truth measurements from oximetry show a clear discrepancy between black and white subjects. In particular, black subjects have higher average blood oxygen. This is compatible with past findings that pulse oximeters overestimate blood oxygen in dark-skinned subjects~\cite{feiner2007dark}. In contrast, the breathing-based oxygen prediction corrects or reduces this bias, and shows more similar oxygen distributions for the two races. 

\section{Extra Results}\label{sec:extra-results}

\subsection{Example of Same Breathing Rate but Different Oxygen Level}
\begin{figure}[hbt]
\centering
\includegraphics[width=0.9\columnwidth]{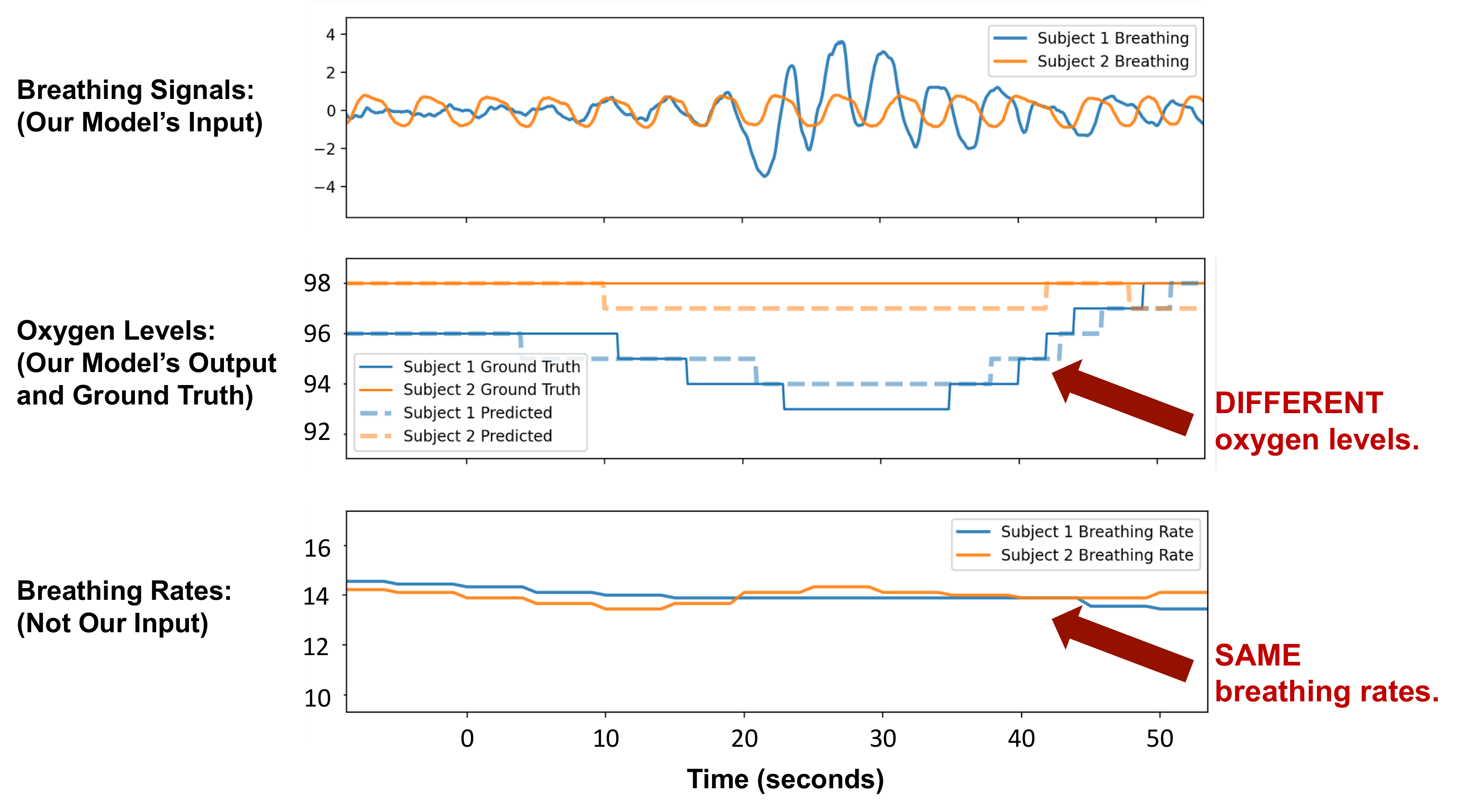}
\vspace{-4mm}
\caption{Example of same breathing rate but different oxygen saturation.}
\label{fig:same-breath}
\end{figure}

One might wonder whether our model predicts oxygen merely by the breathing frequency. We show it is not the case. Our model understands that the same breathing frequency does not translate to the same oxygen level. \Figref{fig:same-breath} shows two people breathing at the same frequency of 14 BPM, but the model correctly realizes that one of them has high and stable oxygen of 98 while the other has a relatively low oxygen level of 93. The model even follows the dynamic change in their oxygen levels from one second to the next. This is possible because the model analyzes the full details of the breathing signal and its dynamics, which is a dense 1D input, not just one variable like the rhythm. To see that, consider again the example in \Figref{fig:same-breath}. While the two people breathe at the same frequency, the person whose breathing is plotted in blue suddenly starts taking deeper breaths. This is an indication that the person is low on oxygen and is trying to increase his oxygen intake. When the person takes such deep breaths, the oxygen level increases. In contrast, the person whose oxygen is plotted in orange is not struggling with low oxygen and his breathing is steady. 

\subsection{Qualitative Results in Different Datasets}
\label{sec:visual}
\begin{figure}[t]
\centering
\includegraphics[width=0.86\columnwidth]{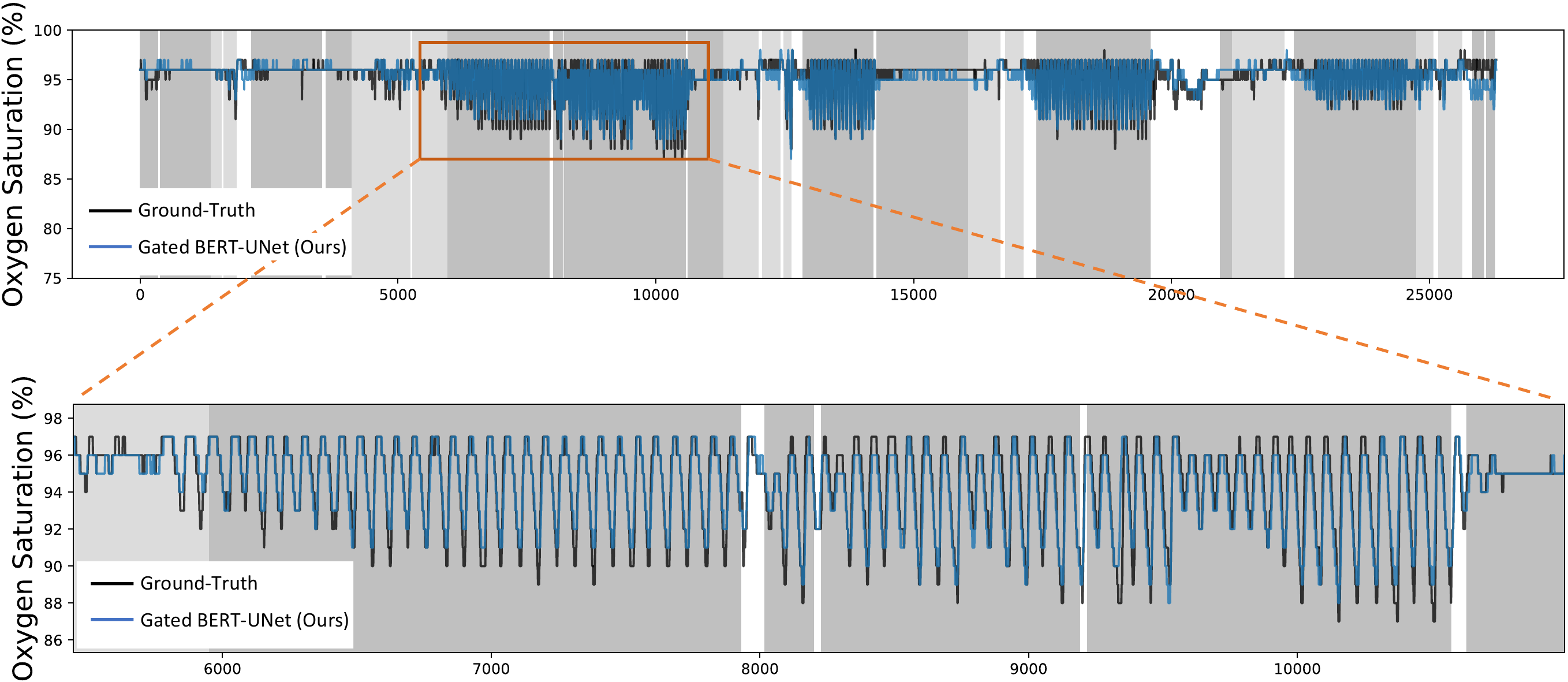}
\caption{\textit{MESA} Example 1.}
\label{fig:mesa1}
\end{figure}

\begin{figure}[h]
\centering
\includegraphics[width=0.86\columnwidth]{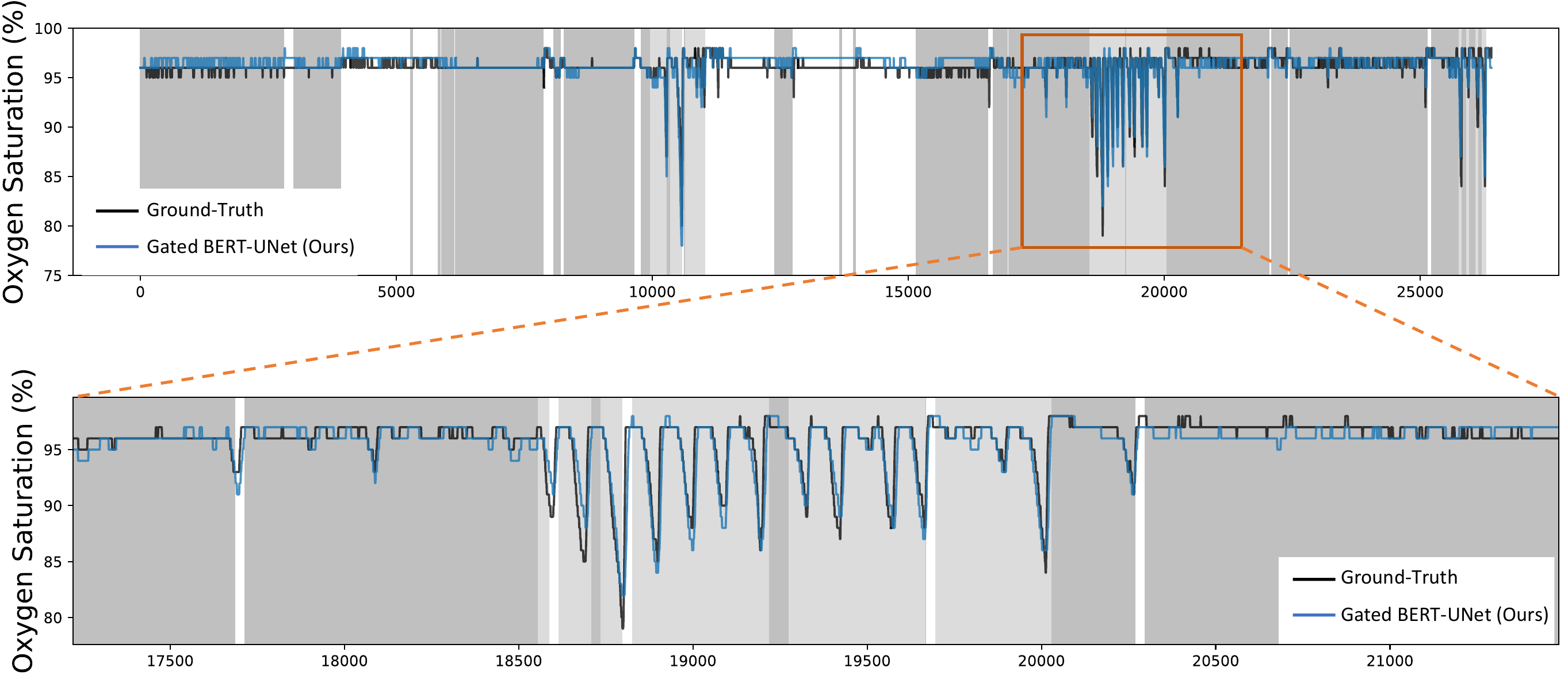} 
\caption{\textit{MESA} Example 2.}
\label{fig:mesa2}
\end{figure}

\begin{figure}[hbt]
\centering
\includegraphics[width=0.86\columnwidth]{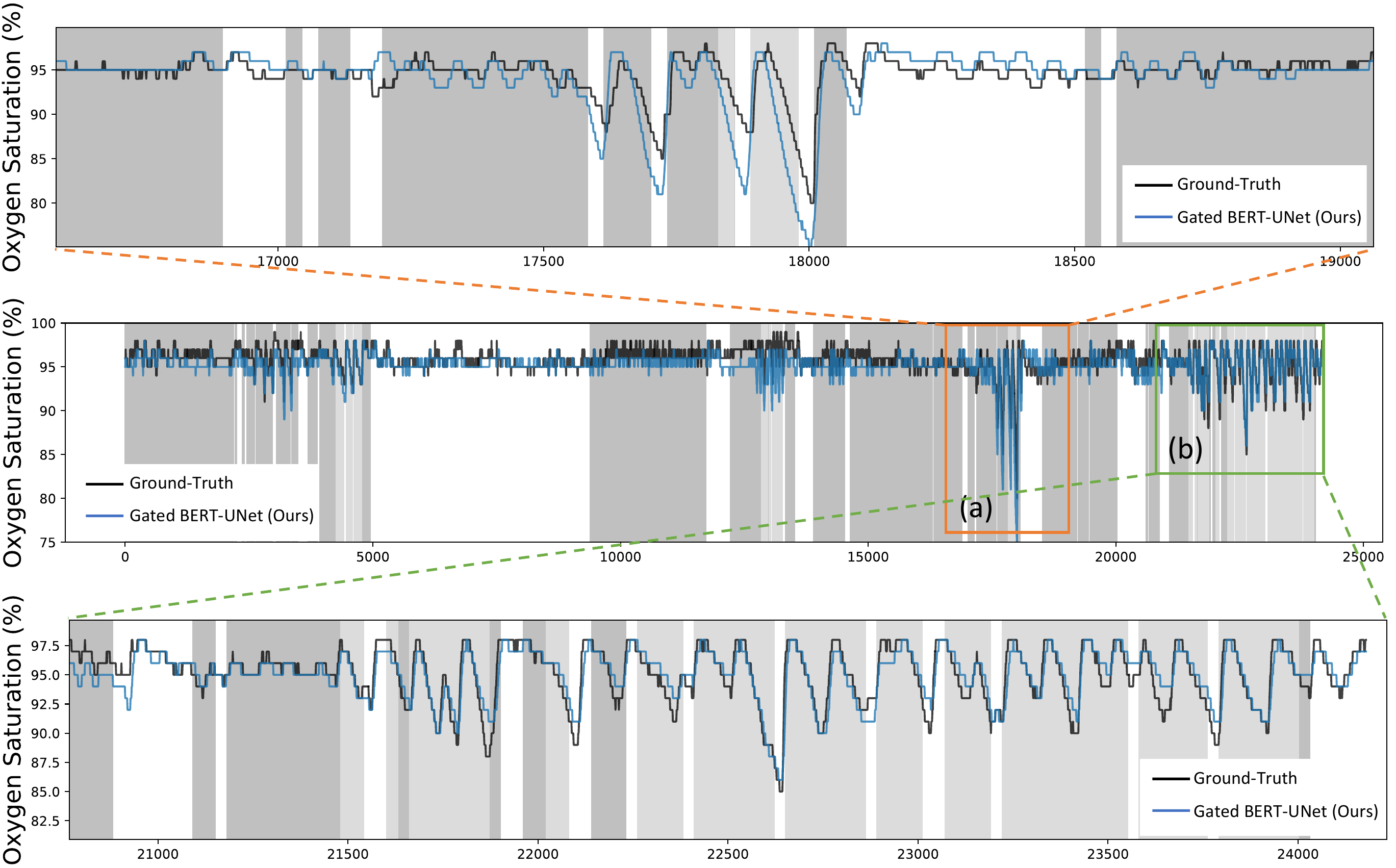}
\caption{\textit{MrOS} Example 1.}
\label{fig:mros1}
\end{figure}

\begin{figure}[hbt]
\centering
\includegraphics[width=0.86\columnwidth]{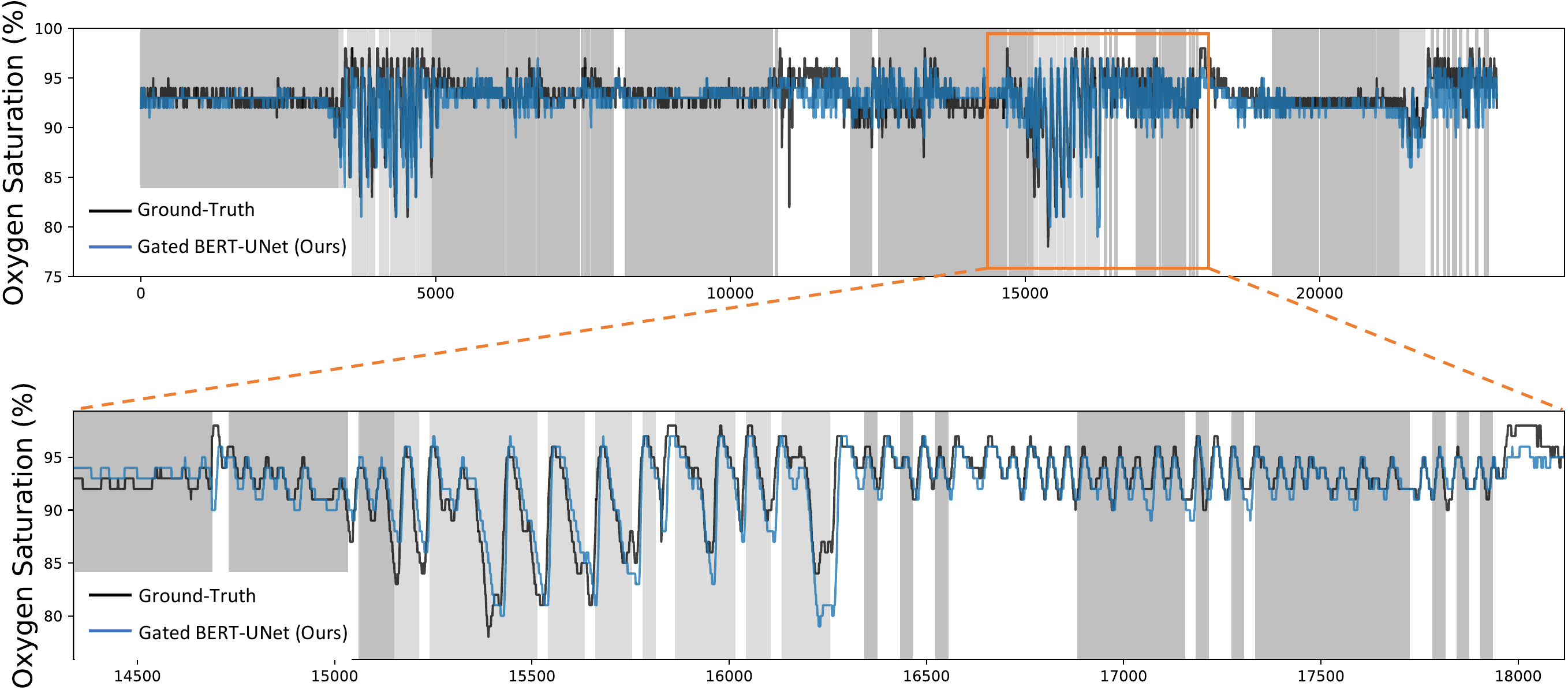}
\caption{\textit{MrOS} Example 2.}
\label{fig:mros2}
\end{figure}

We include more visualizations of the breathing signals and the corresponding oxygen saturation predicted by the \textit{Gated BERT-UNet} model for the medical datasets: MESA (\Figref{fig:mesa1} and \Figref{fig:mesa2}), MrOS (\Figref{fig:mros1} and \Figref{fig:mros2}). We also visualize the model's prediction on unhealthy subjects with various diseases including chronic obstructive pulmonary disease (COPD), asthma, diabetes. In the plots, The background color indicates different sleep stages. The `dark grey', `light grey' and `white' correspond to `Non-REM sleep', `REM' and `Awake', respectively. 


\paragraph{MESA.}
The example in \Figref{fig:mesa1} shows our model's ability to capture the fluctuations of oxygen saturation. At the same time, the example in \Figref{fig:mesa2} shows that our model accurately detects the region of low oxygen saturation, which highlights its usefulness in monitoring patients. 

\paragraph{MrOS.}
From the zoomed-in regions (a) and (b) in \Figref{fig:mros1}, we see that the model exhibits a larger error when the ground truth SpO2 reading is very low. It is mainly caused by the imbalanced labels in the training set since subjects usually experience much less time of having low oxygen level (e.g., below 90\%) than having a normal oxygen level between 94\% to 100\%. \Figref{fig:mros2} shows another example of the dynamics. As shown in the zoomed range, oxygen fluctuations correlate with one's sleep stage: in `REM' (colored by light gray), the oxygen fluctuates drastically while in `Non-REM sleep' (colored by dark gray), the oxygen is much more stable and fluctuates in a small range. Our model makes accurate predictions since it leverages the sleep stage information.

\newpage
\subsection{Results for Relevant Diseases}
As we mentioned earlier, using breathing as the input to the  neural network model allows us to both train and test on large respiration dataset from past sleep studies. Since these datasets contain diverse people with a variety of diseases, it makes it possible to check how the model generalizes to unhealthy individuals. Particularly we are interested in diseases that interact with oxygen saturation including pulmonary diseases such as chronic obstructive pulmonary disease (COPD), chronic bronchitis, asthma, emphysema, and others like diabetes and coronary heart disease.

\begin{figure}[t]
\centering
\includegraphics[width=0.86\columnwidth]{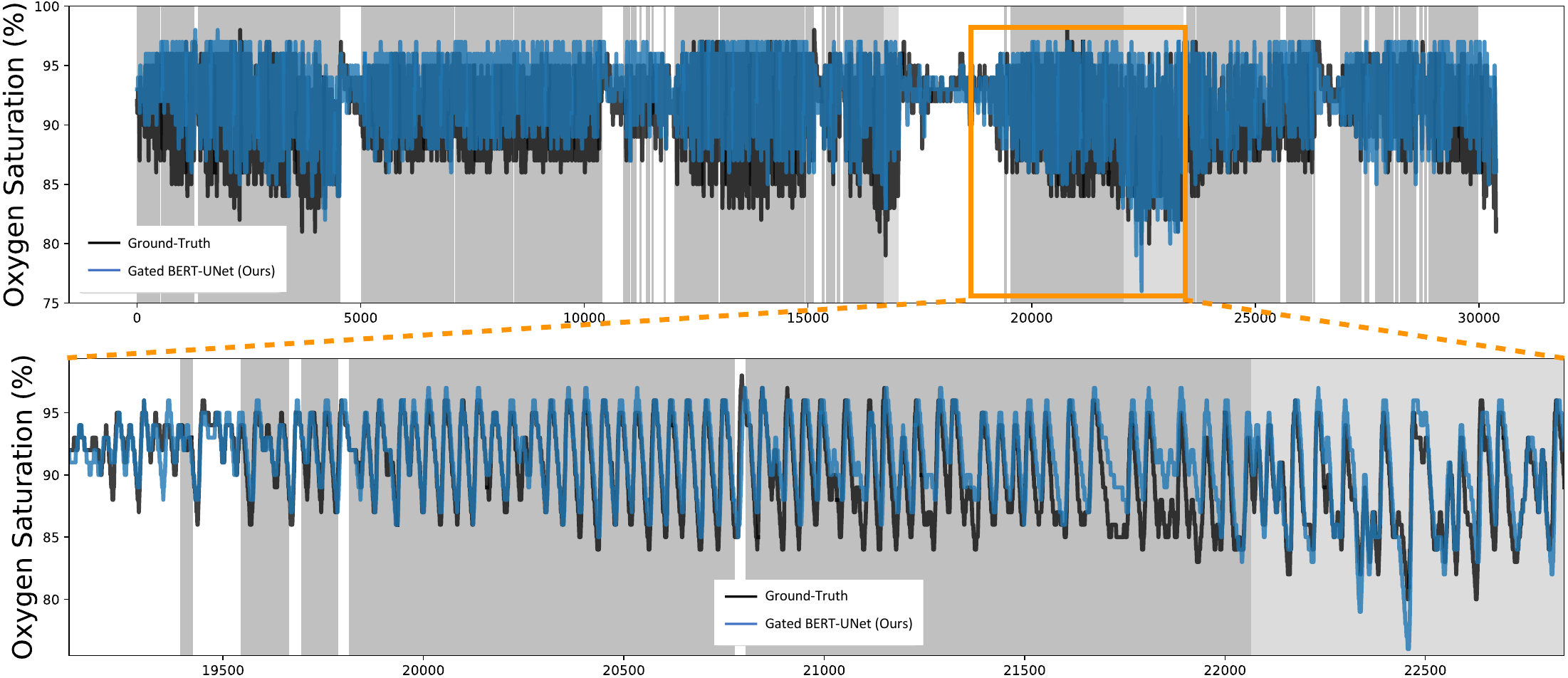}
\caption{Diabetes Patient Example. The first row shows the full-night oxygen saturation, while the second row zooms into the first row's orange box region. 
The black and blue curves are the ground-truth oxygen and our prediction. The background color indicates the subject's sleep stages. The dark grey, light grey and white corresponds to Sleep, REM and Awake respectively.}
\label{fig:diabetes2}
\end{figure}

Diabetes is a disease in which the patient's blood sugar levels are too high. Research has shown that diabetes is a risk factor for severe nocturnal Hypoxemia in obese patients~\cite{lecube2009diabetes}. Further diabetic patients tend to have 3\% to 10\% lower lung volumes than adults without the disease. \Figref{fig:diabetes2} shows an example of a diabetic patient who has an oxygen level that keeps oscillating between 85\% and 95\%. From the zoomed-in region, we can see our model captures the oscillating oxygen dynamics and accurately predicts the oxygen values. 

\begin{figure}[t]
\centering
\includegraphics[width=0.86\columnwidth]{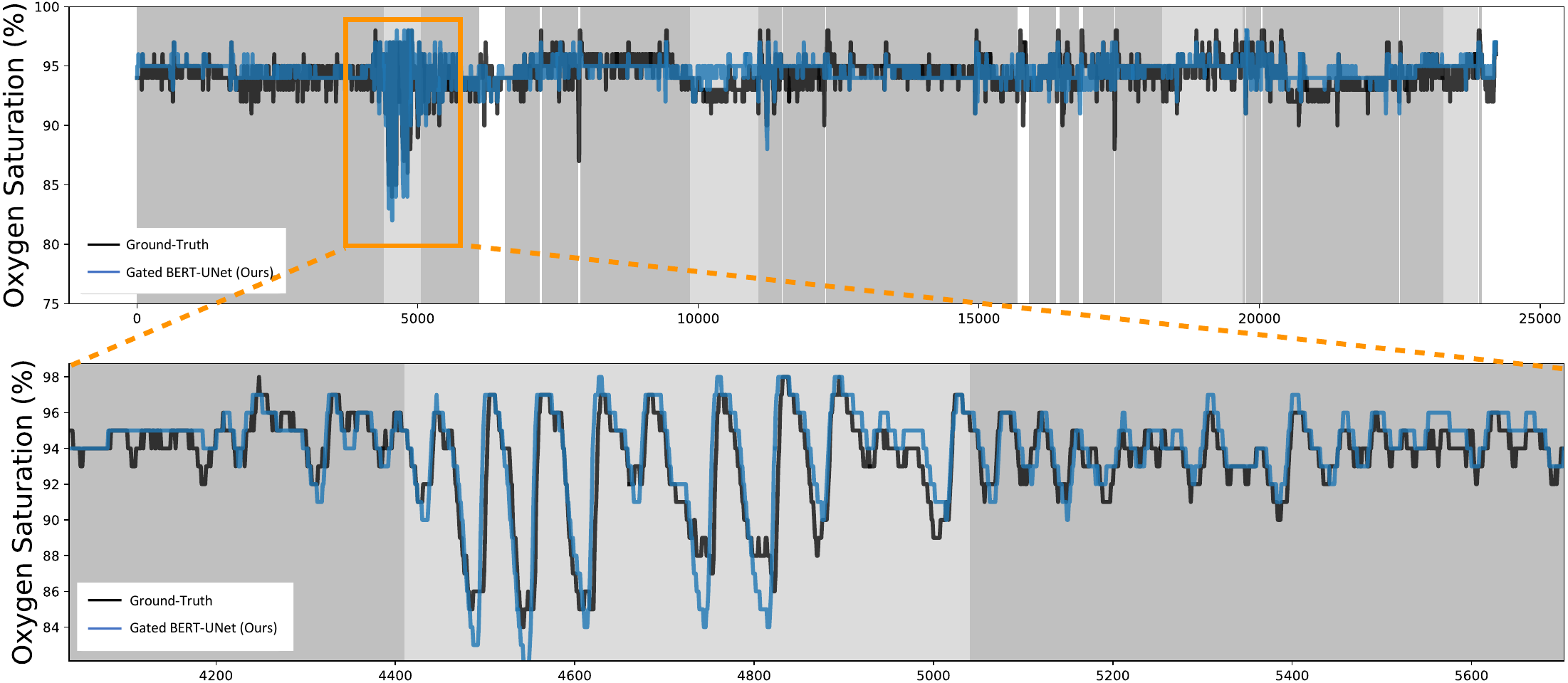}
\caption{Chronic Obstructive Pulmonary Disease (COPD) Patient Example. The first row shows the full-night oxygen level, while the second row zooms into the first row's orange box region. The black and blue curves are the ground-truth oxygen and our prediction. The background color indicates the subject's sleep stages. The dark grey, light grey and white means Sleep, REM and Awake.}
\label{fig:copd1}
\vspace{-4mm}
\end{figure}

Chronic obstructive pulmonary disease (COPD) refers to a chronic inflammatory lung disease that obstructs airflow from the lungs. Severe COPD can cause hypoxia, an extremely low oxygen level. \Figref{fig:copd1} shows an example of a COPD patient who experiences several oxygen droppings during the REM period (indicated by the orange box). As we can see, our model successfully predicts the events of oxygen dropping. 

\begin{figure}[t]
\centering
\includegraphics[width=0.86\columnwidth]{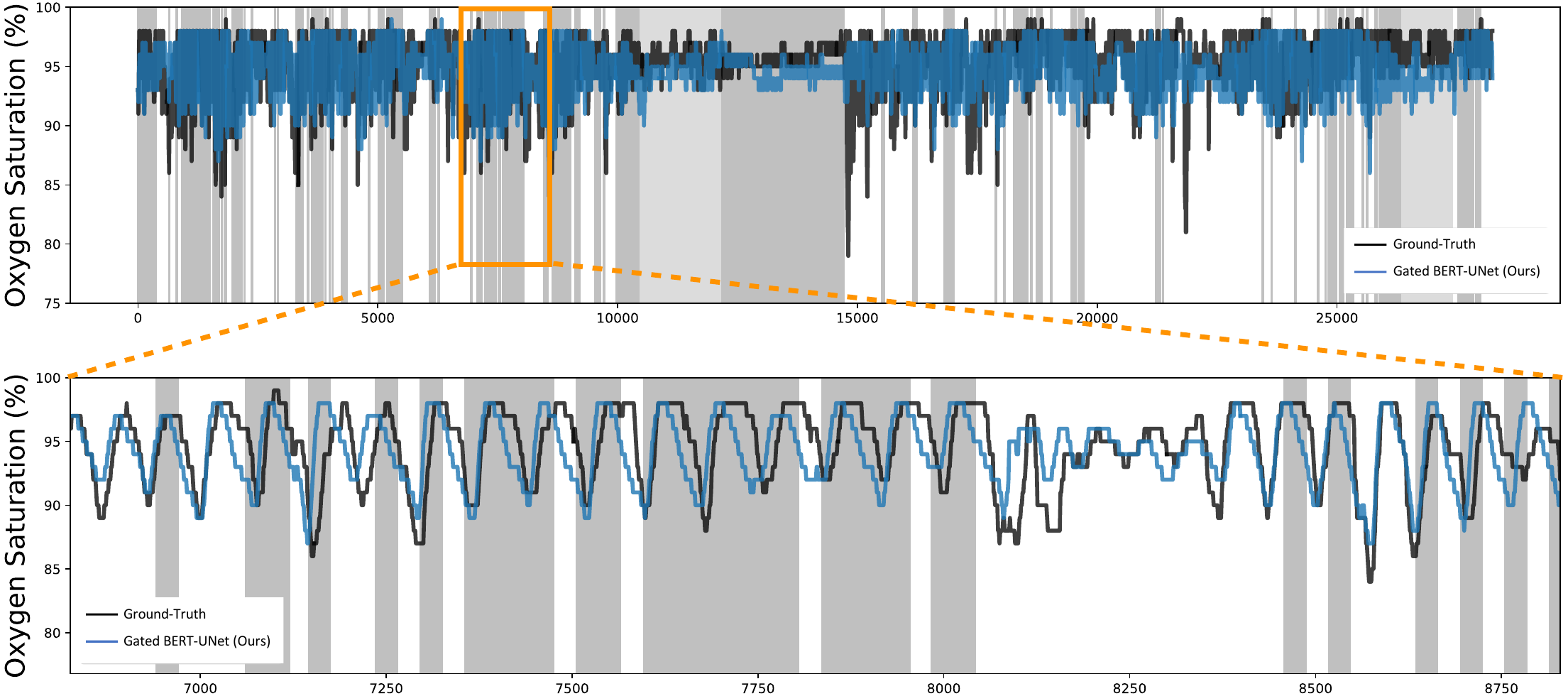}
\caption{Chronic Bronchitis Patient Example. The first row shows the full-night oxygen saturation, while the second row is a zoomed-in visualization of the orange box region in the first row. 
The black and blue curves are the ground-truth oxygen and our prediction. The background color indicates the subject's sleep stages. The dark grey, light grey and white means Sleep, REM and Awake.}
\label{fig:cb2}
\vspace{-4mm}
\end{figure}

Chronic bronchitis refers to long-term inflammation of the bronchi. Chronic bronchitis patients can have shortness of breath which affects oxygen levels. \Figref{fig:cb2} is an example of a chronic bronchitis patient whose oxygen level keeps osculating between normal and low during the night. Our model captures the trend well. 

\begin{figure}[t]
\centering
\includegraphics[width=0.86\columnwidth]{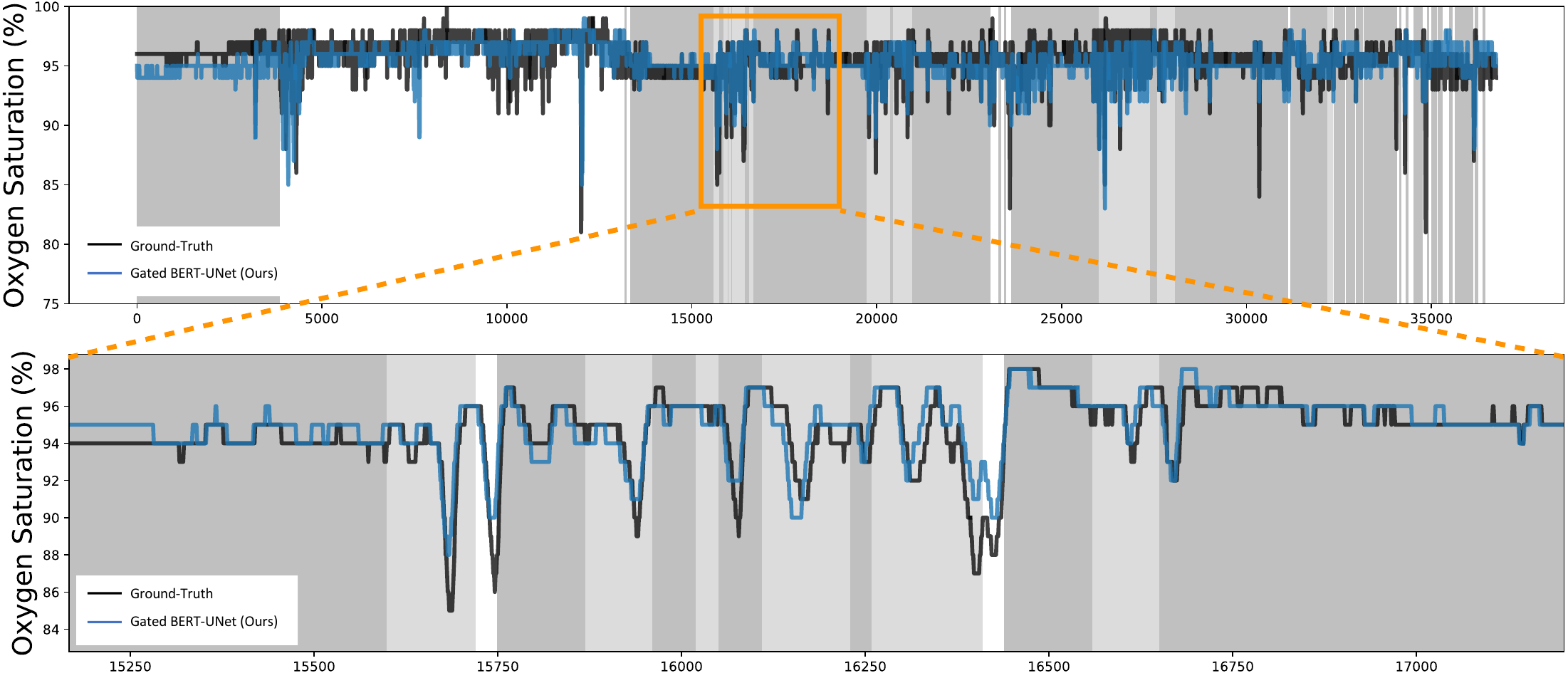}
\caption{Asthma Patient Example.  The first row shows the full-night oxygen saturation, while the second row is a zoomed-in visualization of the orange box region in the first row. 
The black and blue curves are the ground-truth oxygen and our prediction. The background color indicates the subject's sleep stages. The dark grey, light grey and white means Sleep, REM and Awake.}
\label{fig:asthma1}
\vspace{-4mm}
\end{figure}

Asthma is a condition in which a person's airways narrow, swell, and produce extra mucus. An asthma patient's oxygen levels can be irregular due to the breathing difficulty caused by the disease. As shown in the example in \Figref{fig:asthma1}, the patient has a normal oxygen level for most of the time, but the oxygen occasionally drops to low levels. Our model works well on detecting those abnormal oxygen levels from the person's respiration. 

\begin{figure}[h]
\centering
\includegraphics[width=0.86\columnwidth]{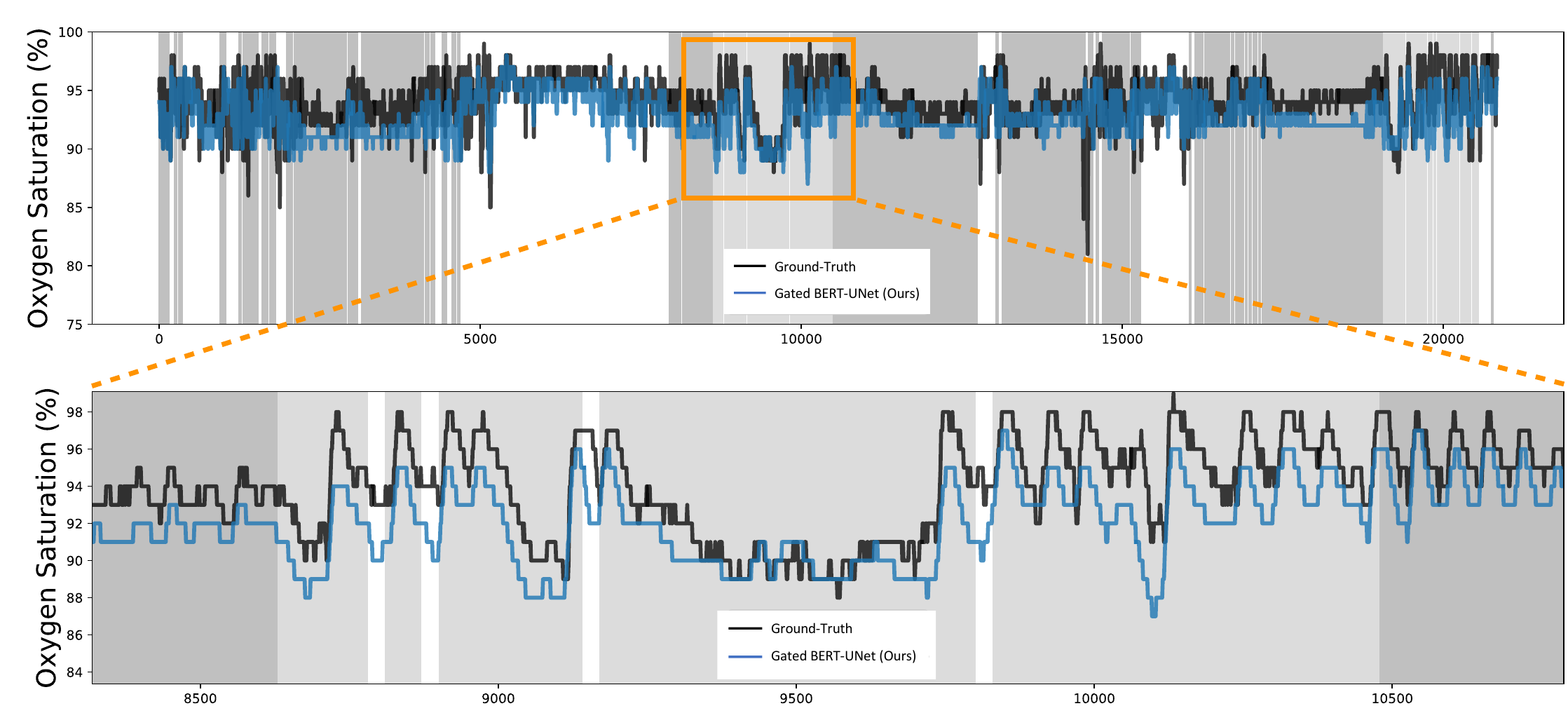}
\caption{Emphysema Patient Example. The first row shows the full-night oxygen saturation, while the second row is a zoomed-in visualization of the orange box region in the first row. 
The black and blue curves are the ground-truth oxygen and our prediction. The background color indicates the subject's sleep stages. The dark grey, light grey and white means Sleep, REM and Awake.}
\label{fig:emphysema2}
\vspace{-4mm}
\end{figure}

Emphysema refers to a lung condition in which the air sacs in the person's lung are damaged. Patients with emphysema usually have breathing issues that affect oxygen saturation. \Figref{fig:emphysema2} shows an example of an emphysema patient. The patient experiences a long period of low oxygen during sleep (as highlighted by the orange box). Our model successfully predicts such unusual oxygen dynamics. 

\begin{figure}[h]
\centering
\includegraphics[width=0.86\columnwidth]{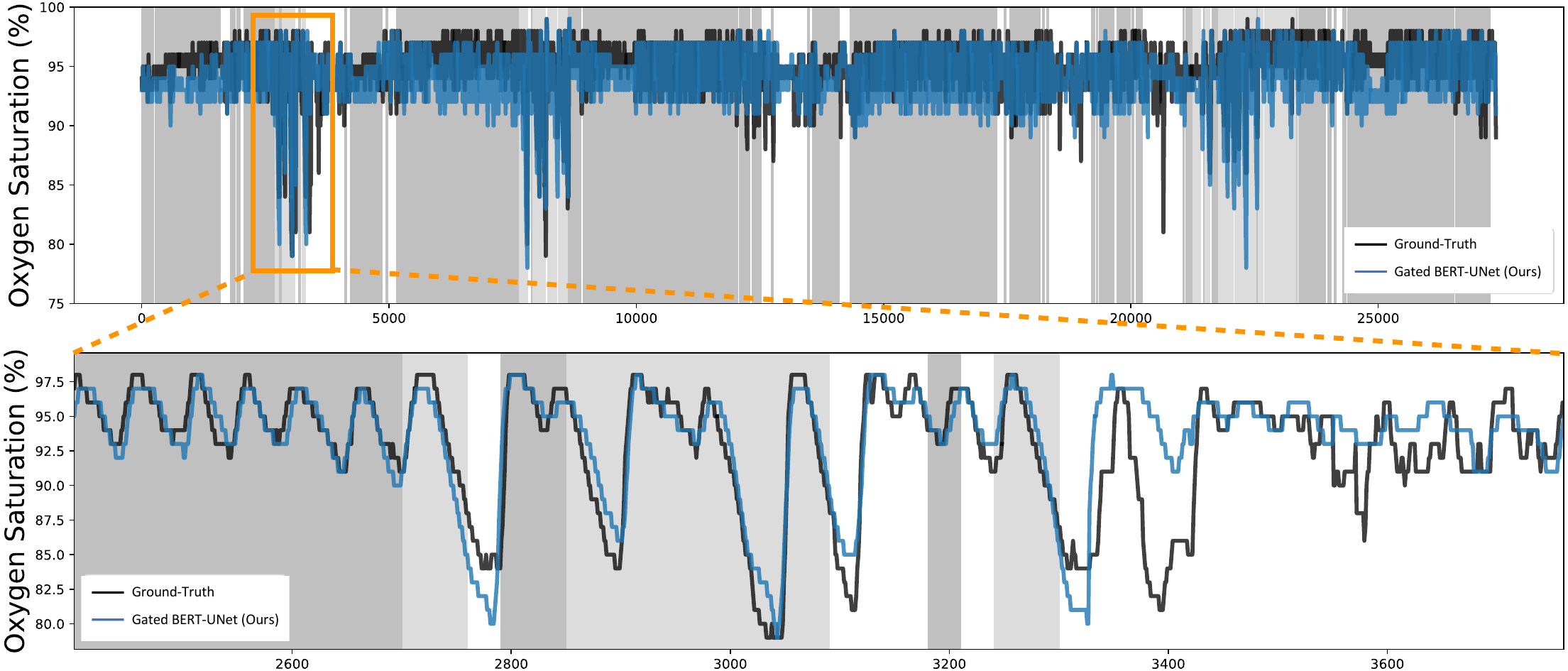}
\caption{Coronary Heart Disease Patient Example. The first row is the full-night oxygen saturation prediction results, where the black curve indicates the ground-truth oxygen saturation and the blue curve indicates our predicted oxygen saturation. The second row is the zoom-in version of the orange box region in the first row. The background color indicates sleep stages. The dark grey, light grey and white corresponds to non-REM Sleep, REM and Awake, respectively.}
\label{fig:chd2}
\vspace{-4mm}
\end{figure}

Coronary heart disease develops when the arteries of the heart are too narrow to deliver enough oxygen-rich blood to the heart. A deficiency in providing oxygen-rich blood can lead one's oxygen saturation to deviate from normal level. \Figref{fig:chd2} is an example of a person with coronary heart disease who experiences several severe oxygen drops during sleep. Our model accurately tracks their oxygen level and detects the oxygen reduction events. 


\newpage

\newpage

\end{document}